\documentclass{ecai}  

\usepackage{graphicx}
\usepackage{latexsym}

\usepackage{multirow}
\usepackage{xcolor}         
\usepackage{colortbl}         
\definecolor{lightgreen}{RGB}{217,255,179}
\usepackage{soul}
\usepackage[ruled,vlined,linesnumbered]{algorithm2e}

\ecaisubmission      

\newcommand{\notesr}[1]{\textcolor{orange}{SR: #1}}

\newcommand{\tblu}[1]{\textcolor{blue}{ #1 }}
\newcommand{\noteng}[1]{\textcolor{red}{\bf \small [#1 --NG]}}

\newcommand{\genemask}{\textsc{GeneMask}}
\newcommand{\genemaskbest}{\textsc{GeneMaskBest}}

\begin{document}

\begin{frontmatter}

\title{\genemask: Fast Pretraining of Gene Sequences to Enable Few-Shot Learning}

\if{0}
\author[A]{\fnms{First}~\snm{Author}\orcid{....-....-....-....}\thanks{Corresponding Author. Email: somename@university.edu.}}
\author[B]{\fnms{Second}~\snm{Author}\orcid{....-....-....-....}}
\author[B]{\fnms{Third}~\snm{Author}\orcid{....-....-....-....}} 

\address[A]{Short Affiliation of First Author}
\address[B]{Short Affiliation of Second Author and Third Author}
\fi

\author[A]{\fnms{Soumyadeep}~\snm{Roy}\thanks{Corresponding Author. Email: soumyadeep.roy9@iitkgp.ac.in The paper is accepted for publication at the 26th European Conference on Artificial Intelligence ECAI 2023} }
\author[B]{\fnms{Jonas}~\snm{Wallat}}
\author[B]{\fnms{Sowmya S}~\snm{Sundaram}}
\author[B]{\fnms{Wolfgang}~\snm{Nejdl}}
\author[A]{\fnms{Niloy}~\snm{Ganguly}}

\address[A]{Indian Institute of Technology Kharagpur}
\address[B]{L3S Research Center, Germany}

\begin{abstract}%
Large-scale language models such as DNABert and LOGO aim to learn optimal gene representations and are trained on the entire \textit{Human Reference Genome}. However, standard tokenization schemes involve a simple sliding window of tokens like \textit{k-mers} that do not leverage any gene-based semantics and thus may lead to (trivial) masking of easily predictable sequences, and subsequently inefficient Masked Language Modeling (MLM) training. Therefore, we propose a novel masking algorithm, \genemask, for MLM training of gene sequences, where we randomly identify positions in a gene sequence as mask centers and locally select the span around the mask center with the highest Normalized Pointwise Mutual Information (NPMI) to mask. We observe that in the absence of human-understandable semantics in the genomics domain (in contrast, semantic units like words and phrases are inherently available in NLP), \genemask-based models substantially outperform the SOTA models (DNABert and LOGO) over four benchmark gene sequence classification datasets in five few-shot settings (10 to 1000-shot). More significantly, the \genemask-based DNABert model is trained for less than one-tenth of the number of epochs of the original SOTA model. We also observe a strong correlation between top-ranked PMI tokens and \textit{conserved DNA sequence motifs}, which may indicate the incorporation of latent genomic information. The codes (including trained models) and datasets are made publicly available at \url{https://github.com/roysoumya/GeneMask}.

\end{abstract}

\end{frontmatter}

\section{Introduction}

Computational analysis of genomics has revolutionized the field of medical science \cite{McGuire2020}, particularly with the advent of the \textit{Human Reference Genome} \cite{Schneider072116}. As seen in recent studies~\cite{enformer,genomics-survey}, deep learning has been applied to various genomic applications, such as protein structure analysis, gene expression data, and transcriptome analysis.
The state-of-the-art pretrained models (DNABert~\cite{DNABert} and LOGO~\cite{yang:2021:bioarxiv:logo}) in gene sequence classification tasks are widely used in literature~\cite{Badirli2021,genebert2021}. The input data for these tasks are often presented as a sequence of nucleotides. Each side of the double-helix DNA strand comprises the bases adenine (A), cytosine (C), guanine (G), and thymine (T). Similar to the NLP domain, the standard approach in gene sequence classification is to pretrain the transformer models by randomly masking (and predicting) tokens, the so-called masked language modeling objective (MLM). However, unlike words or sentences in languages, no clear semantically demarcated tokens are present within the gene sequence. Therefore, to come to a workable solution, researchers randomly select a sequence of $k$ nucleotides~\cite{DNABert,genebert2021}; for example, if $k$ = 3, the sequence $TCG$ can be selected from a  gene sequence  $\cdots GAT\textbf{TCG}ATGC\cdots$. 
However, the workable solution may be easy to guess if (say) GAT\textbf{TCG} is a very common sequence because \textbf{GAT} is still unmasked and frequently co-occurs with the masked \textbf{TCG} sequence. This, in turn, may decelerate the training process and increase the pretraining time; 
for example, DNABert trains for $25$ days on $8$ NVIDIA 2080Ti GPUs. A possible solution to reduce the instances of `easy' learning is to identify highly-correlated commonly occurring spans and mask them in their entirety so that the pretraining model does not consume precious computing cycles in predicting `easy' cases.

A systematic approach to identify correlated spans can be based on the principle of \textit{Pointwise Mutual Information} (PMI), which calculates the chance of a set of tokens spanning together vis-a-vis occurring independently; a high PMI score indicates a high correlation. In this work, we propose a novel masking algorithm, \genemask, for MLM training of gene sequences. We randomly identify positions in a gene sequence as mask centers and locally select the span around the mask center with the highest (modified) PMI to mask.  
However, PMI also favors rarely occurring correlated spans  (where individual tokens may have a low frequency of occurrence), which are not easy cases as the model has not frequently seen such patterns.
Hence we modify the PMI metric and impose a discounting factor to down-score rare correlated spans. 
Instead of randomly selecting tokens to mask, as done in DNABert (or LOGO), we use the PMI score to prioritize the \textit{relevant} tokens within a given gene sequence. In the case of \genemask, we randomly chose positions in a gene sequence and locally chose the span with the highest PMI to mask. This is in contrast to the strategy adopted in NLP literature by Levine et al.~\cite{pmi-masking}, where they use the PMI score to formulate an absolute importance notion and consequently create a fixed masking vocabulary. 

As current medical datasets often face data scarcity issues, the move towards personalized medicine requires models that perform well with limited training data at hand \cite{shaikhina:2017:aimed:SmallDataMLinMedicine,Hekler:2019:bmcmedicine:smallDataParadigm}. We, therefore, evaluate our proposed \genemask-based MLM training strategy in the low-resource (few-shot) setting. 

Our extensive experimentation shows that \genemask-based DNABert and LOGO improve over the standard random masking-guided DNABert and LOGO, respectively, in \textbf{few-shot settings} ($10, 50, 100, 500$ and $1000$ training data points per class) over four benchmark datasets of gene sequence classification (two on promoters - \textit{Prom-core, Prom-300}, one on enhancer - \textit{Cohn-enh} and one on splice sites - \textit{Splice-40}). We posit that \genemask{} helps incorporate non-trivial genetic knowledge because we observe that \genemask-based DNABert pretrained for \textbf{10K steps even outperforms original DNABert pretrained for 120K steps} for all few-shot settings in case of Prom-300 and Cohn-enh dataset. 
In addition, we perform \textbf{motif}\footnote{Sequence motifs are short, recurring patterns in DNA that are presumed to have a biological function~\cite{haeseleer2006}.}
\textbf{analysis} and observe a strong correlation between top-ranked PMI tokens and conserved DNA sequence motifs, providing a biological reason behind the performance improvement in \genemask-based DNABert. 
Finally, to alleviate the issue of the tremendous engineering effort needed to develop the experimental setup of gene sequence classification tasks, we make all the codes (including trained models), data, and appendix publicly available at \url{https://github.com/roysoumya/GeneMask}.
\section{Background}\label{sec:prior-art}

\noindent \textbf{Learning deep representations in the context of gene sequence modeling.}
Recent studies~\cite{dnavec,DNABert,yang:2021:bioarxiv:logo} represent gene sequences as k-mers, using a learned dense representation from an adapted BERT model. Mock et al.~\cite{mock2021bertax} broadly adopt the DNABert architecture for the task of taxonomy classification and model gene sequences as 3-mers and also include next sentence prediction along with MLM training loss. 
 Instead of using k-mer representations, the \textit{BigBird} model~\cite{bigbird2020} trained a SentencePiece tokenizer on the Human Reference Genome and applied to the tasks of chromatin-profile prediction and promoter region prediction. 
However, the pretrained model weights of the \textit{BigBird} model for the genomics setting are not publicly available, and the alternative of training from scratch is prohibitively costly for such a large model.
Mo et al.~\cite{genebert2021} infuse domain knowledge into the model by proposing a multimodal pretraining setup comprising gene sequences and information on transcription factors and regions. The recent Enformer~\cite{enformer} utilizes a combination of convolutional and transformer layers to feed long sequences of one-hot encoded base pairs into the model. However, unlike gene sequence classification, which is the focus of this work, the Enformer model is used to predict gene expression tracks. 

\noindent \textbf{Random and PMI-masking for MLM training in NLP.} 
The initial work applied \emph{random token masking}, as performed by BERT \cite{bert}, where 15\% of the input tokens are chosen to be masked uniformly. Previous work has also investigated simultaneous masking of sequences of adjacent tokens which form either a whole word (\emph{whole word masking} \cite{senrich:2016:acl:wholewordmasking}) or an entity (\emph{entity masking} \cite{sun2019ernie}); the technique proves to be beneficial over randomly masking single, non-contiguous tokens. 
Joshi et al.~\cite{spanbert2020} propose \emph{random span masking} where random spans with lengths chosen from a geometric distribution are masked at random positions. This simple method outperforms the more involved entity masking approach~\cite{sun2019ernie}. When applied to gene sequences, especially entity and whole word masking, have the problem that there is no well-defined concept of entities or words in gene sequences. 
The \emph{PMI-masking} approach \cite{pmi-masking} builds on the ideas of the entity and span masking where they treat collocated n-grams with high Pointwise Mutual Information scores analogous to an entity and thereby mask these spans of tokens together. The authors show that masking PMI tokens (i) accelerates training while matching end-of-pretraining performance in roughly half the training steps and (ii) improves upon previous masking approaches at the end of pretraining. Sadeq et al.~\cite{sadeq-etal-2022-informask} develop the \textit{InforMask} masking strategy, where they measure the \textit{informative relevance} of a word as the sum of PMI values between a masked word and all unmasked words in the given sentence (higher values are prioritized for masking). However, the semantic equivalent of a sentence (in NLP) is not known in the case of gene sequence modeling.
\section{Building blocks of SOTA models}\label{sec:dnabert-config}

We focus on the two SOTA transformer-based pretrained models named DNABert~\cite{DNABert} and LOGO~\cite{yang:2021:bioarxiv:logo} that are adapted to the gene sequence modeling domain. Through MLM pretraining, these models learn powerful contextual representations for DNA fragments utilizing abundant unlabeled data from the \textit{Human Reference Genome}, which contains around 3.2 billion base pairs over 24 chromosomes. 

\noindent \textbf{Implementation details of SOTA models and associated research challenges.} 
The tokenization of gene sequences and MLM training is performed based on the author's codebase~\cite{DNABert-Codebase}. 
However, they do not provide the dataset for pretraining or finetuning (downstream) tasks. Therefore, we follow the author's description to construct the corresponding datasets, which is nontrivial. We explain the pretraining data creation process in this section and later describe the finetuning dataset creation process in Section~\ref{sec:datasets}. 

\noindent \textbf{Tokenization of gene sequences.} The gene sequence is first converted into a k-mer representation, which is commonly used in the literature~\cite{DNABert,dnavec}. The $k$-mer representation is a sliding window of length $k$. For example, \textit{AGCACGCAG} in 6-mer representation leads to 3 tokens - \textit{AGCACG}, \textit{GCACGA}, \textit{CACGAG}. The vocabulary comprises all combinations ($4^k$ length) and five special tokens - \textit{CLS, PAD, UNK, SEP, MASK}. According to the set-up chosen by SOTA models, we consider $k = 6$ for all the experiments, as it also provides a good trade-off between longer contextual information and manageable computational complexity~\cite{yang:2021:bioarxiv:logo}. 
As stated in the LOGO paper~\cite{yang:2021:bioarxiv:logo}, 6-mers incorporate richer contextual information while keeping the memory and computational complexity manageable.  
\begin{figure*}[!ht]
\centering
\includegraphics[width=0.65\textwidth]{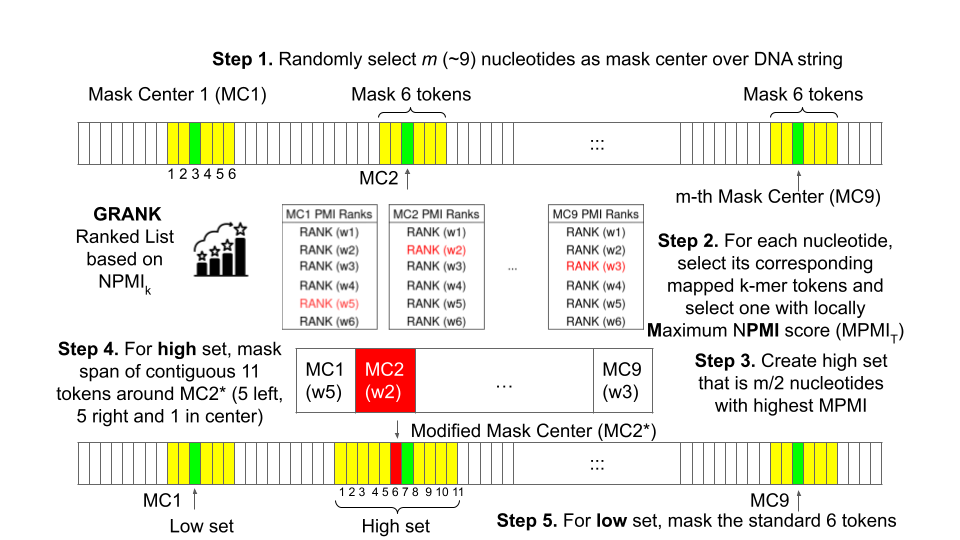}
\caption{ Method overview of \genemask, our proposed masking algorithm for MLM training of gene sequences, where we randomly select mask centers over the input DNA string and locally select the span around the mask center with the highest Normalized Pointwise Mutual Information (NPMI\textsubscript{k}) to mask }
\label{fig:method-overview}
\end{figure*}

\noindent \textbf{Pretraining data preparation.} 
We obtain the Human Reference Genome from the Genome Reference Consortium Human Build 38 patch release 13 (GRCh38.p13) FASTA file~\cite{human-reference} from the NCBI website. 
It serves as a large-scale corpus of unlabeled gene sequence data, which we use for MLM training to obtain a contextual representation of the 6-mer tokens. We perform the following steps to convert the Human Reference Genome to a form that DNABert (or LOGO) can use to train with the MLM objective: \\
    {\bf (a).} For each chromosome $c$ in the Human Reference Genome, we randomly choose the starting index between $1$ and $1000$~\cite{bigbird2020}. \\
    {\bf (b).} Given the chromosome number and its starting index ($ST$), we next determine the length of the DNA segment $L$ as BERT has the limitation of accommodating a maximum of $512$ tokens. We select $L$ as $510$ for $50\%$ of the cases and a randomly selected length between $5$ and $510$ for the remaining $50\%$ of cases~\cite{DNABert}. \\
    {\bf (c).} We thus create a DNA segment comprising the base pairs between $ST$ and $ST+L$ of chromosome $c$, corresponding to \textit{data point in the pretraining dataset}. We filter out DNA segments that contain bases other than A, T, C, or G. 


\noindent \textbf{MLM training.} The SOTA models are trained with masked language modeling loss similar to  BERT~\cite{bert}. 
However, to mask a nucleotide, a contiguous sequence of tokens is masked to prevent information leakage, as each nucleotide is part of k consecutive k-mers. 
More formally, say a nucleotide is represented as DNA[$i$], while a 6-mer token is represented as T[$i$], equivalent to \{DNA[$i$-2], DNA[$i$ - 1] $\cdots$ DNA[$i$ + 3]\}; then the tokens T[$j$],  $\forall(j)_{j=i - 2}^{i + 3}$ are masked. We enforce this mapping between a nucleotide and a 6-mer token in \genemask{} as the $MapNucleotideToKmerTokens$ function in Algorithm~\ref{algo:genemask}.
Given that, $k=6$ and $15\%$ of tokens need to be masked~\cite{bert}, the MLM probability is set at: $15\% / 6$; that is, $2.5\%$ of the nucleotides are chosen for masking. (Since we are mainly working with 6-mer, unless mentioned explicitly T[$i$] = $T_6[i]$.)

\if{0}
DNABert is a BERT-based transformer model that is trained on gene sequences obtained from the Human Reference Genome, which corresponds to a long sequence of nucleotides (a combination of bases adenine (A), cytosine (C), guanine (G), and thymine (T)) for each chromosome. We use the Genome Reference Consortium Human Build 38 patch release 13 (GRCh38.p13) FASTA file with RefSeq assembly accession id as GCF\_000001405.39 from NCBI website\footnote{\url{https://www.ncbi.nlm.nih.gov/assembly/GCF_000001405.39/}} for constructing the large gene sequence corpus to be used to perform unsupervised Masked Language Model (MLM) training. It is presented as a sequence of nucleotides. Each side of the double helix DNA strand comprises the bases adenine (A), cytosine (C), guanine (G), and thymine (T)).

We convert the obtained human reference genome into documents ($D$), where each document $d \in D$ is a sequence of sentences, where each sentence is a sequence of fragments of DNA.

DNABert first takes a set of sequences represented as k-mer tokens as input. 

Each sequence is then represented as a matrix M by embedding each token into a numerical vector. Specifically, DNABert encodes contextual information by performing the multi-head self attention mechanism. It adopts similar pretraining scheme like BERT with the following key differences: (a) next sentence prediction task is not used for pretraining, therefore only the masked language modeling loss remains (b) since each token is in k-mer representation, thus the previous token has ($k-1$) base pairs overlapping. Thus, we always mask a contiguous sequence of $k$ tokens while masking in order to prevent overfitting. 

\noindent \textbf{Pretraining Data Construction.} \tblu{ Following the pretraining data preparation strategy as outlined in DNABert, we obtain sequences of length 510 for 50\% of cases and a randomly selected length between 5 and 510 for the remaining half of the time.} 
 
\noindent \textbf{Tokenization of gene sequences.} \noteng{I think we dont have this. }\notesr{Removed gene boundary description}
The gene sequence is first converted into a k-mer representation, which is commonly used in literature~\cite{dnavec,DNABert}. ~\cite{Yang2021} further supported such a k-mer representation from a biological standpoint, stating that each nucleotide is not independent such as codon rules in coding region and regulatory motifs in non-coding region. $k$-mer representation is a sliding window of length $k$. For example, \textit{AGCACGCAG} in 6-mer representation leads to 3 tokens - \textit{AGCACG}, \textit{GCACGA}, and \textit{CACGCAG}. The vocabulary thus comprises of all combinations ($4^k$ length) and 5 special tokens - \textit{CLS, PAD, UNK, SEP, MASK}. We consider $k = 6$ for all the experiments, and interchangeably refer to it as $6-mer$.
\fi
\section{Methodology}\label{sec:method}
\genemask{} aims to mask the k (=6)-mers that co-occur much more than expected compared to their components (i.e., k-mers of shorter length, such as 4-mers or 5-mers). These spans then replace the (uniform) random masking strategy used by SOTA models. Removing highly correlated local contexts make the masked token prediction task more difficult, which may improve the pretraining efficiency~\cite{pmi-masking}. 
However, \genemask{} significantly differs from the \textit{PMI-masking} strategy~\cite{pmi-masking} used in NLP, which first uses the (unnormalized) PMI score to formulate an absolute importance notion and subsequently creates a fixed masking vocabulary. To the best of our knowledge, this is the first work to develop a principled (statistical) approach to identify a highly correlated span of tokens in gene sequences based on normalized PMI score (Section~\ref{sec:pmiscoring}) and then develop a novel masking algorithm (\genemask) for efficient MLM training of gene-based (pretrained) SOTA models like DNABert and LOGO (Section~\ref{sec:entitydnabert}); Figure~\ref{fig:method-overview} provides the methodology overview.

\subsection{PMI-based Metric to Identify Correlated Spans in Gene Sequences}\label{sec:pmiscoring}

This work considers a single nucleotide equivalent to a single token in NLP. Thus an n-gram from the NLP domain is equivalent to a k-mer from the gene sequence modeling literature.
We propose a novel strategy to adapt PMI-based scoring to our genomic setting, which can help us identify high PMI tokens to mask. \textit{Pointwise Mutual Information} (PMI) quantifies how often two tokens occur compared to what is expected if they are independent. The PMI formula (proposed by Levine et al.~\cite{pmi-masking}) when extended to k-mers (where k $>$ 2) is:
\begin{equation}\label{eq:pmi}
	\textrm{PMI}_k(w_1\ldots w_k)=\min_{\sigma\in\textrm{seg}(w_1\ldots w_k)}\log\frac{p(w_1\ldots w_k)}{\prod_{s\in\sigma}p(s)}
	\end{equation}
\raggedbottom
Here, $\textrm{seg}(w_1\ldots w_k)$ is the set of all contiguous segmentations of the $k$-mer ``$w_1\ldots w_k$" (excluding the identity segmentation). In a valid segmentation ($\sigma$), the original sequence ``$w_1\ldots w_k$" can be divided into any number of partitions of positive ($>0$) size. For example, say for $k=6$, some of the possible valid segmentations are: ``($w_1\ldots w_3$), ($w_4\ldots w_6$)" or ``($w_1,w_2$) ($w_3$), ($w_4\ldots w_6$)".

\textbf{The PMI\textsubscript{k} formulation many times favors tokens with lower frequency}, that is, the number of times the k-mer gene sequence appears in the \textit{Human Reference Genome}. We thus impose a \textbf{discounting factor} that penalizes rare tokens~\cite{pantel2002} because such PMI tokens will be more frequently selected as masked tokens during the MLM training stage. This may lead the model to over-emphasize corner cases and thus wrongfully mask `not-so-easy' cases. We refer to it as the \textit{Normalized PMI\textsubscript{k}} (NPMI\textsubscript{k}) formula, which we finally use for scoring all the individual n-gram sequences. 

{\small{
\begin{equation}\label{eq:norm-pmi}
	\textrm{NPMI}_k(w_1\ldots w_k)= \newline PMI\textsubscript{k} * \frac{\log f(w_1\ldots w_k)}{\log (c)+ \log f(w_1\ldots w_k)} 
	\end{equation}
}}
\raggedbottom
Here, $f(w_1\ldots w_k)$ refers to the frequency of occurrence of the k-mer sequence of $w_1\ldots w_k$. $c$ refers to the minimum frequency of occurrence (a constant value used as a threshold to remove rare tokens). In this paper, we focus only on computing NPMI\textsubscript{k} for all k-mer sequences where $k = 6$, and develop {\bf a ranked list (\textit{GRANK}) of all 6-mers} ($4096$ in total) based on the decreasing order of NPMI\textsubscript{k}. Next, we discuss how we use the NPMI\textsubscript{k} scores as a  measure to choose tokens to be masked during MLM training.

\subsection{Masking Algorithm for Efficient Pretraining over Gene Sequences} \label{sec:entitydnabert}
 \genemask{} aims to mask all the nucleotides simultaneously in the most correlated spans. Masking correlated spans minimizes information leakage and helps the system to learn deeper patterns. Side by side, we would like to preserve the benefit drawn out of the traditional random masking strategy. Therefore, we propose a novel linear-time masking algorithm, \genemask, for MLM training of gene sequences, where we randomly identify positions in a gene sequence as mask centers and locally select the span around the mask center with the highest NPMI\textsubscript{k} to mask. We present the details of our proposed \genemask{} strategy in Algorithm~\ref{algo:genemask}. 

 \newcommand\mycommfont[1]{\footnotesize\ttfamily\textcolor{blue}{#1}}
\SetCommentSty{mycommfont}
\SetKwInput{KwInput}{Input}      
\SetKwInput{KwOutput}{Output}    
\SetKwInput{KwData}{Initialization}

\LinesNumberedHidden

\begin{algorithm}[ht]
\footnotesize
\caption{\genemask{} Algorithm}\label{algo:genemask}

\DontPrintSemicolon \;
    \KwInput{DNA string made of 6-mer tokens with a maximum length of 510, Dictionary containing NPMI\textsubscript{k} values for all 6-mers }
    \KwOutput{$MaskTokenSet$: List of tokens to mask in the DNA string}
    \KwData{
    \tcp{T[i] denotes a 6-mer token at $i$-th position in the DNA segment, DNA[i] denotes the $i$-th nucleotide}

    $MaskTokenSet \gets \emptyset$ 
    
    T[$i$] $\gets$ \{DNA[$i$-2]~$\cdots$~DNA[$i$+3]\}
    }
	\SetKwFunction{FMain}{MapNucleotideToKmerTokens}
	\SetKwProg{Fn}{Function}{:}{}
	
	\Fn{\FMain{nucleotide position id $i$}}{

            $MappedTokens$ $\gets$ T[$j$],  $\forall(j)_{j=i - 2}^{i + 3}$ \;
			
			\KwRet\  $MappedTokens$ \; 
    }

    \textbf{Step 1:} Randomly select $m$ nucleotides as mask centers (MC) spread uniformly over the DNA string.
            
            \tcp{In this step corresponding to each mask center, the neighboring nucleotide  which locally has the highest NPMI\textsubscript{k} is chosen}
    \textbf{Step 2:}
    \For{each nucleotide in $m$ mask centers}
	{
            $PositionId$ := Token index of nucleotide on the DNA segment given as input

            $KmerTokens$ := $MapNucleotideToKmerTokens$ ($PositionId$)

            \tcp{Select kmer with the highest $NPMI_k$ score, T[$\tau$] }
            
            $T[\tau]$ $\gets \arg$ $\max_{kmer \in KmerTokens}$ $NPMI_k$( kmer )

            \tcp{Store the Locally Maximum NPMI score as MPMI\textsubscript{$\tau$}}
            \textit{MPMI\textsubscript{$\tau$}} $\gets$  NPMI\textsubscript{k} $(T[\tau])$
            
	} 

     \textbf{Step 3:} Divide the $m$ nucleotides into two sets based upon their \textit{MPMI} scores, where the high set is the $m/2$ nucleotides with the highest \textit{MPMI} scores. 

     \textbf{Step 4:} 
     
    \For{each nucleotide in high set}	{
        \tcp{Mask all the nucleotides in T[$\tau$] (this ensures the masking of correlated spans together, 11 tokens in length)}
        $MaskTokenSet$ $\gets$ $MaskTokenSet$ $~\cup$ $\forall(j)_{j=\tau - 2}^{\tau + 3}$ $MapNucleotideToKmerTokens$( j )
     }

     \For{each nucleotide in low set}{
            \tcp{Mask only the corresponding nucleotide DNA[$i$] (this mimics a random masking strategy)}
        $MaskTokenSet \gets MaskTokenSet \cup MapNucleotideToKmerTokens( i )$
     }

     \KwRet\  $MaskTokenSet$ \;
\end{algorithm}

\noindent {\bf Determining the value of \textit{m}:} To mask six base pairs-long gene sequence that is all the nucleotides in a particular token T[$i$], we need to mask a span of contiguous $11$ tokens (6 mask centers, two tokens to the left and three tokens right) while as mentioned a single nucleotide induce masking of $6$ tokens. Thus, the expected mask span length per mask center is computed as $0.5 * 6 + 0.5 * 11 = 8.5$. Subsequently, the \textit{mlm probability} is updated from $2.5\%$ to $1.765\%$ (= $15\% / 8.5$). Hence when the DNA length is $512$, $m$ $\approx$ 9. 

\noindent \textbf{Efficiency of \genemask{} algorithm.} \genemask{} first samples $m$ random nucleotides as mask centers and locally chooses the one with the highest NPMI\textsubscript{k} score within the span (fixed number of adjacent tokens) of the mask center. Thus the time complexity of \genemask{} becomes $O(m)$. Since $m = c*n$, where $n$ is the input sequence length and $c$ is a constant value (equal to $1.765\%$ in our case), the time complexity of \genemask{} is: $O(m)$ = $O (c*n)$ = $O(n)$. 

\section{Experimental Setup}\label{sec:experiments}
Here, we provide the dataset specifics, evaluation setup,  model training, and baseline model details.

\subsection{Datasets}\label{sec:datasets}
We use four benchmark datasets of gene sequence classification {(\small{\textsc{Prom-core, Prom-300, Splice-40, Cohn-enh}})} for evaluation purposes. The two datasets of promoter region prediction and splice site prediction are not directly  available and involve significant effort (including a paper implementation) for their construction. 

\noindent {\small{\textbf{\textsc{Promoter Region Prediction (Prom-core and Prom-300).}}}}
A \textit{promoter} is a DNA region typically located upstream of the gene, which is the site of transcription initiation (as defined by Zaheer et al.~\cite{bigbird2020}). 
\ul{The task is to classify a given DNA fragment as a promoter or non-promoter sequence}. 
 However, we follow the instructions of the DeePromoter~\cite{deepromoter2019} paper, including the negative data creation, since the authors do not provide the datasets. Thus, we obtained human TATA and non-TATA promoter data, i.e., including promoter sequences with and without a TATA box (a common promoter-related motif found between $-30$ to $-25$ bp (upstream) of a gene's transcription start site), from the Eukaryotic Promoter Database~\cite{epdnew2012}, using the website API of the \textit{EPD
selection tool}~\cite{EPD}. 
We extracted $-249$ to $+50$ bp sequences around TSS for the {\bf Prom-300 dataset} and $-34$ to $+35$ bp for the {\bf Prom-core dataset.} We perform the standard train-test split of $70\%$ and $30\%$, which leads to $53276$ and $5920$ data points, respectively. 

\noindent {\small{\textbf{\textsc{Splice Donor and Acceptor Site Prediction (Splice-40).}}}}
We followed the same strategy as prior works~\cite{DNABert,Wang2019} for dataset construction. We extract $40$~bp long sequences around the donor and acceptor sites of exons (randomly selected) as positive sequences. 
The exon's position information (chromosome number, start index, and end index) is obtained from the corresponding gene annotation file. \ul{The task is to classify a given DNA fragment as a donor, acceptor, or non-splice site (non-overlapping intermediate sequences between exons) sequence.} We perform the standard train-test split of $70-30\%$, which leads to $24300$ and $3000$ training and test data points, respectively. 

\noindent {\small{\textbf{\textsc{Enhancer Cohn Prediction (Cohn-enh).}}}}
An \textit{enhancer} is a DNA sequence that can bind specific proteins and increase the chance of transcription of a particular gene. \ul{Here, the input is a DNA sequence of $500$ bp in length and a binary classification task where the task is to classify a DNA fragment as an enhancer or non-enhancer sequence.}
This dataset has been adapted from Cohn et al.~\cite{cohn2018} and is made available as a benchmark dataset by Martinek et al.~\cite{martinek2022} in Github~\cite{gresova2022genomic}; we use the same train-test split that leads to $20843$ and $6948$ data points as train and test datasets, respectively. 

\subsection{Evaluation Setup}\label{sec:evaluation-setup}
We report the standard metrics of accuracy (used for performance comparison) and AUC (stands for Area Under the Receiver Operating Characteristic Curve) used for classification tasks where the class labels are balanced. 
We follow the standard evaluation setup used in the few-shot text classification setting~\cite{schick-schutze-2021-exploiting,schick-schutze-2021-just}. Thus, we assume not to have access to a validation dataset to optimize the hyperparameters and investigate the performance for different training set sizes (few-shot settings) $n= 10, 50, 100, 500,$ and $1000$, where $n$ denotes the number of training data points per class. 
We report the mean and standard deviation of accuracy and AUC by running the experiments \textbf{ten} times by randomly choosing seed and $n$ training (fine-tuning) data points per class in each run.
We report the statistical significance results based on paired t-test for the performance improvement by \genemaskbest{} over ORI 10K model. 

\subsection{Training Details}\label{sec:impl-details}

\begin{figure}[t]
\centering
\includegraphics[width=\columnwidth]{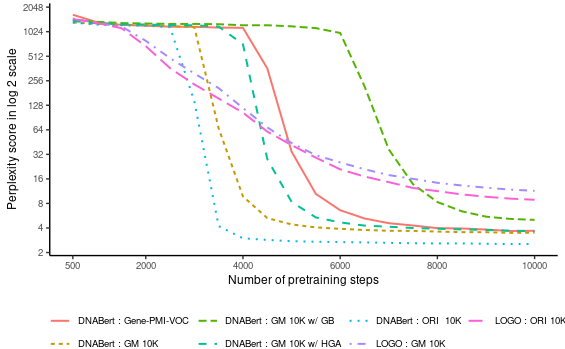}
\caption{Perplexity score plot of \genemask{} (GM)}
\label{fig:perplexity-plot}
\end{figure}

In this paper, we train all the pretrained models and their variants for only $10000$ steps, which takes about \textbf{2.5 days to complete for DNABert} and around \textbf{20 hours for LOGO} using four GTX 1080Ti 11GB GPUs.  We select such a setup for two reasons --- (i) to explore different pretrained model variants in a reasonable time because we observe that the perplexity of DNABert (SOTA model) has converged to a low score and is stable over the last $3000$ pretraining steps (see Figure~\ref{fig:perplexity-plot}).
(ii) As observed by Levine et al.~\cite{pmi-masking}, PMI masking learns fast and is thus quite efficient to reap the benefit even with fewer pretraining steps. The default number of warmup steps for DNABert is 10K steps out of 200K ($5\%$ of the maximum number of steps). Since we reduce the maximum pretraining steps limit from 200K to 10K, we set the number of warmup steps as $500$ ($5\%$ of $10000$ steps) to maintain the same ratio. 
We use the same values as the original SOTA models for the remaining hyperparameters 
(see section~\ref{sec:appendex-model-parameter} of Appendix to know more about training (including fine-tuning) details and hyperparameters). We observe in Figure~\ref{fig:perplexity-plot} that the perplexity scores of the baseline models are lower.  The same trend is also reported by Levine et al.~\cite{pmi-masking}. However, the single-token perplexity values are not comparable between models with different masking strategies and, thus, do not indicate downstream performance.

\subsection{Baseline Models}\label{sec:baseline-models}
We evaluate our work on two gene transformer-based models - DNABert~\cite{DNABert} and LOGO~\cite{yang:2021:bioarxiv:logo}. To perform a fair comparison, we train the baseline models with the same hyperparameter settings as \genemask. These models follow \textbf{random masking} during the MLM training step instead of the proposed \genemask{} strategy.  We use the same model hyperparameters for pretraining \genemask{} and the baseline models to perform a fair comparison. 
 We will refer to the SOTA models that use random masking as the original SOTA model \textbf{(ORI)} without any PMI-guided masking.  We use the original DNABert model pretrained on 120K steps (\textbf{ORI 120K}) based on the pretrained model weights provided by ~\cite{DNABert} as a baseline model. 
 
\noindent\textbf{Fixed PMI-guided Masking Vocabulary (Gene-PMI-VOC)}: We undertook a significant effort to adapt the PMI-masking strategy in NLP proposed by Levine et al.~\cite{pmi-masking} to the genomic setting which is a contribution in itself. We create a PMI-masking vocabulary $\approx 10$ times the DNABert vocabulary size of $4101$ tokens. 
 We first select all possible k-mer sequences ($2 \leq k \leq 10$) whose frequency of occurrence is  $\geq 10000$. We then rank them using our proposed PMI metric and select the top $40000$ as the masking vocabulary. During masking, we randomly select the mask centers (nucleotides) and then use the masking vocabulary to tokenize the input gene sequence into PMI tokens. We mask an entire PMI token (2 to 10-mer) within which the selected mask center lies (based on tokenization); Algorithm~\ref{algo:genepmivoc} (appendix) describes our implementation in detail. 

 \noindent \textbf{Pretraining Model Variants of \genemask.} We explore two different pretraining setups --- (i) Half Gradient Accumulation (task-independent) and (ii) Gene Boundary-aware Pretraining (task-specific, boosts performance over Splice-40 dataset).

    \noindent \textbf{(i). Half Gradient Accumulation (HGA)}: The \textit{gradient accumulation steps}  parameter is halved, reducing it from $25$ (default DNABert configuration) to $12$ steps, and consequently, it reduces the effective batch size (it is the product of per GPU train batch size, GPU count and gradient accumulation steps) by $50\%$. This aims to reduce the \textit{generalization gap} issue that arises when the training batch size is too large~\cite{hoffer2017,keskar2017}. 
    
   \noindent \textbf{(ii). Gene Boundary-aware Pretraining (GB)}: In the pretraining data construction stage, as previously described in Section~\ref{sec:dnabert-config}, we do not consider gene boundaries. Thus, it may happen that the DNA segment spans across different genomic entities and, since such spans have no semantic underpinning, may introduce noise during the masked token prediction step of MLM training. We obtain the gene boundary information from the gene annotation file, corresponding to the same Human Reference Genome, as described in the pretraining data construction part of Section~\ref{sec:dnabert-config}. To limit a DNA segment within a single gene boundary, we use the minimum distance to the next gene boundary as an upper limit for determining the length of the DNA segment $L$; each DNA segment corresponds to a single data point in the pretraining dataset. 

\section{Experimental Results}
We compare the performance of our proposed \genemask{} strategy and random masking strategy-based SOTA models of DNABert and LOGO as shown in Table~\ref{tab:perf-compare-10K}. We observe that the performance improvement due to \genemask{} strategy over random masking strategy (ORI 10K model) is more prevalent in lower data settings (10, 50-shot), reduces at higher data settings of 500 and 1000-shot (except for Splice-40), and ceases to exist when trained on full datasets, as evident from Figure~\ref{fig:perf-compare-graph} of Appendix. We consider the two additional pretraining model variations for DNABert - 
\textit{Half Gradient Accumulation} and \textit{Gene Boundary-aware Pretraining} and along with the base model referring to the \genemask{} model having the same hyperparameters as the baseline models such as ORI 10K. We present the best performance among the three models, such as \genemask{}, \genemask{w/ HGA} and \genemask{w/ GB}] as \genemaskbest{} model. The \genemask{} model is also referred to as \genemask{} 10K to indicate the number of training steps. 

\begin{table*}[t]
    \footnotesize
    \centering
    \scalebox{0.8}{
    \addtolength{\tabcolsep}{-0.1em}
    \begin{tabular}{cccccccccc}
     \hline
        \textbf{Data} &  \textbf{Model}  & \multicolumn{2}{c}{\textbf{Prom-core}}& \multicolumn{2}{c}{\textbf{Prom-300}} & \multicolumn{2}{c}{\textbf{Cohn-enh}}& \multicolumn{2}{c}{\textbf{Splice-40}}   \\
        \textbf{per class}&\textbf{type}&\textbf{Accuracy}&\textbf{AUC}&\textbf{Accuracy}&\textbf{AUC}&\textbf{Accuracy}&\textbf{AUC}&\textbf{Accuracy}&\textbf{AUC} \\ \hline 
        &\multicolumn{9}{c}{\textbf{DNABert as Base model}} \\ 
        10& ORI 120K &\textbf{0.606 $\pm$ 0.045}&\textbf{0.661 $\pm$ 0.064}&$0.638 \pm 0.070$&$0.708 \pm 0.088$&$0.582 \pm 0.030$&$0.631 \pm 0.044$&$0.404 \pm 0.019$&$0.577 \pm 0.021$ \\
        
& Gene-PMI-VOC &$0.6 \pm 0.047$&$0.652 \pm 0.067$&\cellcolor{yellow} $0.653 \pm 0.065$&\cellcolor{yellow} $0.731 \pm 0.066$&\cellcolor{yellow} $0.596 \pm 0.048$& \cellcolor{yellow} $0.66 \pm 0.062$ &\textbf{0.412 $\pm$ 0.02} &\textbf{0.599 $\pm$ 0.02} \\
        &  ORI 10K &$0.586 \pm 0.051$&$0.641 \pm 0.082$& $0.601 \pm 0.065$ &$0.657 \pm 0.095$&$0.579 \pm 0.047$& $0.655 \pm 0.058$ &$0.409 \pm 0.017$&$0.588 \pm 0.018$\\
 
&  \cellcolor{lightgreen} \genemaskbest{} &\cellcolor{yellow} 0.602 $\pm$ 0.058&\cellcolor{yellow} 0.655 $\pm$ 0.078&\cellcolor{lightgreen} \textbf{0.676 $\pm$ 0.054}**&\cellcolor{lightgreen} \textbf{0.779 $\pm$ 0.074}**&\cellcolor{lightgreen} \textbf{0.622 $\pm$ 0.050}*&\cellcolor{lightgreen} \textbf{0.701 $\pm$ 0.038}* &\cellcolor{lightgreen} \textbf{0.412 $\pm$ 0.026}&\cellcolor{yellow} $0.591 \pm 0.036$\\ \cline{2-10}

50& ORI 120K &\textbf{0.687 $\pm$ 0.024}&\textbf{0.756 $\pm$ 0.03}&\cellcolor{yellow} $0.808 \pm 0.019$&$0.89 \pm 0.013$&$0.638 \pm 0.020$&$0.679 \pm 0.028$ &$0.472 \pm 0.048$&$0.651 \pm 0.049$ \\ 
& Gene-PMI-VOC  &$0.649 \pm 0.058$&$0.738 \pm 0.033$&$0.800 \pm 0.027$&\cellcolor{yellow} $0.893 \pm 0.02$&\cellcolor{yellow} $0.645 \pm 0.014$&\cellcolor{yellow} $0.706 \pm 0.016$ &\textbf{0.522 $\pm$ 0.071}&\textbf{0.703 $\pm$ 0.062}\\
&  ORI 10K &$0.653 \pm 0.058$&$0.718 \pm 0.064$&$0.789 \pm 0.059$&$0.882 \pm 0.04$&$0.634 \pm 0.031$&$0.689 \pm 0.044$ &$0.512 \pm 0.014$&$0.687 \pm 0.017$\\

&\cellcolor{lightgreen}  \genemaskbest{} &\cellcolor{yellow} 0.678 $\pm$ 0.026&\cellcolor{yellow} 0.744 $\pm$ 0.026&\cellcolor{lightgreen} \textbf{0.815 $\pm$ 0.02}&\cellcolor{lightgreen} \textbf{0.905 $\pm$ 0.013}&\cellcolor{lightgreen} \textbf{0.654 $\pm$ 0.017}*&\cellcolor{lightgreen} \textbf{0.713 $\pm$ 0.011}$\dagger$ &\cellcolor{yellow} $0.519 \pm 0.027$&\cellcolor{yellow} $0.696 \pm 0.023$\\ \cline{2-10}

100& ORI 120K &\textbf{0.712 $\pm$ 0.009}&\textbf{0.781 $\pm$ 0.012}&\cellcolor{yellow} $0.842 \pm 0.014$&$0.915 \pm 0.012$&\cellcolor{yellow} $0.669 \pm 0.017$&$0.736 \pm 0.022$ &$0.507 \pm 0.059$&$0.683 \pm 0.063$\\ 
& Gene-PMI-VOC  &$0.697 \pm 0.011$&$0.767 \pm 0.013$&$0.835 \pm 0.017$&$0.912 \pm 0.014$&$0.65 \pm 0.051$&\textbf{0.737 $\pm$ 0.011} &\textbf{0.605 $\pm$ 0.017}&\textbf{0.779 $\pm$ 0.015}\\ 

&  ORI 10K &$0.695 \pm 0.014$&$0.765 \pm 0.017$&\cellcolor{yellow} $0.842 \pm 0.018$&\textbf{0.923 $\pm$ 0.009}&$0.668 \pm 0.015$&$0.736 \pm 0.011$ &$0.543 \pm 0.027$&$0.72 \pm 0.024$\\

&\cellcolor{lightgreen} \genemaskbest{} &\cellcolor{yellow} 0.708 $\pm$ 0.013*&\cellcolor{yellow} 0.779 $\pm$ 0.015*&\cellcolor{lightgreen} \textbf{0.847 $\pm$ 0.029}& \cellcolor{yellow} 0.920 $\pm$ 0.020&\cellcolor{lightgreen} \textbf{0.67 $\pm$ 0.017}&\cellcolor{lightgreen} \textbf{0.737 $\pm$ 0.013} &\cellcolor{yellow} $0.577 \pm 0.019$**&\cellcolor{yellow} $0.751 \pm 0.017$**\\ \cline{2-10}

500& ORI 120K &$0.743 \pm 0.008$&$0.819 \pm 0.007$&$0.883 \pm 0.006$&$0.951 \pm 0.005$&\textbf{0.698 $\pm$ 0.009}&\textbf{0.776 $\pm$ 0.011} &$0.429 \pm 0.07$&$0.608 \pm 0.077$\\ 

& Gene-PMI-VOC &$0.738 \pm 0.021$&$0.82 \pm 0.017$&$0.884 \pm 0.004$&$0.948 \pm 0.008$&$0.692 \pm 0.011$&$0.759 \pm 0.011$&$0.639 \pm 0.025$&$0.809 \pm 0.021$ \\ 
&  ORI 10K &\cellcolor{yellow} $0.752 \pm 0.007$&\textbf{0.831 $\pm$ 0.003}&\cellcolor{yellow} $0.888 \pm 0.007$&\textbf{0.958 $\pm$ 0.002}&$0.696 \pm 0.009$&$0.767 \pm 0.008$&\cellcolor{yellow} $0.665 \pm 0.035$&\cellcolor{yellow} $0.834 \pm 0.022$ \\

&\cellcolor{lightgreen}  \genemaskbest{} &\cellcolor{lightgreen} \textbf{0.753 $\pm$ 0.005}&\cellcolor{lightgreen} \textbf{0.831 $\pm$ 0.003}&\cellcolor{lightgreen} \textbf{0.89 $\pm$ 0.006}&\cellcolor{yellow} $0.957 \pm 0.002$&\cellcolor{lightgreen} \textbf{0.698 $\pm$ 0.006}&\cellcolor{yellow} 0.771 $\pm$ 0.007 &\cellcolor{lightgreen} \textbf{0.692 $\pm$ 0.016}*&\cellcolor{lightgreen} \textbf{0.851 $\pm$ 0.01}$\dagger$\\ \cline{2-10}

1000& ORI 120K  &$0.758 \pm 0.006$&$0.835 \pm 0.004$&$0.895 \pm 0.005$&$0.957 \pm 0.005$&$0.700 \pm 0.009$&$0.769 \pm 0.009$ &$0.496 \pm 0.109$&$0.673 \pm 0.107$\\ 
& Gene-PMI-VOC &$0.759 \pm 0.007$&$0.834 \pm 0.006$&$0.895 \pm 0.004$&$0.96 \pm 0.004$&$0.698 \pm 0.007$&$0.766 \pm 0.009$ &$0.504 \pm 0.02$&$0.689 \pm 0.016$ \\ 
&ORI 10K &\cellcolor{yellow} $0.765 \pm 0.004$&\cellcolor{yellow} $0.839 \pm 0.005$&\textbf{0.901 $\pm$ 0.003}&\textbf{0.964 $\pm$ 0.002}&\cellcolor{yellow} $0.705 \pm 0.005$&\cellcolor{yellow} $0.776 \pm 0.006$ &\cellcolor{yellow} $0.651 \pm 0.019$&\cellcolor{yellow} $0.821 \pm 0.015$\\

&\cellcolor{lightgreen}  \genemaskbest{} &\cellcolor{lightgreen} \textbf{0.766 $\pm$ 0.007}&\cellcolor{lightgreen} \textbf{0.843 $\pm$ 0.006}&\cellcolor{yellow} $0.898 \pm 0.005$&\cellcolor{yellow} $0.962 \pm 0.002$&\cellcolor{lightgreen} \textbf{0.706 $\pm$ 0.005}&\cellcolor{lightgreen} \textbf{0.778 $\pm$ 0.006} &\cellcolor{lightgreen} \textbf{0.709 $\pm$ 0.012}**&\cellcolor{lightgreen} \textbf{0.865 $\pm$ 0.008}** \\ \hline 

&\multicolumn{9}{c}{\textbf{LOGO as Base model}} \\ 
10& ORI 10K&$0.506 \pm 0.017$&$0.557 \pm 0.064$&$0.502 \pm 0.005$&$0.557 \pm 0.028$&$0.53 \pm 0.043$&$0.599 \pm 0.067$ &$0.333 \pm 0.012$&$0.565 \pm 0.022$\\
&\cellcolor{lightgreen} \genemask{} 10K &\cellcolor{lightgreen} \textbf{0.54 $\pm$ 0.049}$\dagger$&\cellcolor{lightgreen} \textbf{0.582 $\pm$ 0.088}$\dagger$&\cellcolor{lightgreen} \textbf{0.519 $\pm$ 0.04}&\cellcolor{lightgreen} \textbf{0.594 $\pm$ 0.086}&\cellcolor{lightgreen} \textbf{0.553 $\pm$ 0.063}&\cellcolor{lightgreen} \textbf{0.649 $\pm$ 0.067}* &\cellcolor{lightgreen} \textbf{0.353 $\pm$ 0.017}*&\cellcolor{lightgreen} \textbf{0.558 $\pm$ 0.028}\\

50& ORI 10K&$0.565 \pm 0.048$&$0.635 \pm 0.037$&$0.618 \pm 0.037$&$0.663 \pm 0.05$&$0.627 \pm 0.007$&$0.676 \pm 0.006$ &$0.438 \pm 0.019$&$0.61 \pm 0.021$\\
&\cellcolor{lightgreen} \genemask{} 10K &\cellcolor{lightgreen} \textbf{0.611 $\pm$ 0.04}**&\cellcolor{lightgreen} \textbf{0.676 $\pm$ 0.017}**&\cellcolor{lightgreen} \textbf{0.668 $\pm$ 0.014}**&\cellcolor{lightgreen} \textbf{0.733 $\pm$ 0.022}**&\cellcolor{lightgreen} \textbf{0.635 $\pm$ 0.006}$\dagger$&\cellcolor{lightgreen} \textbf{0.692 $\pm$ 0.009}** &\cellcolor{lightgreen} \textbf{0.465 $\pm$ 0.018}**&\cellcolor{lightgreen} \textbf{0.636 $\pm$ 0.022}*\\

100& ORI 10K&$0.628 \pm 0.014$&$0.677 \pm 0.017$&$0.646 \pm 0.033$&$0.694 \pm 0.039$&$0.638 \pm 0.008$&$0.691 \pm $ 0.011&$0.447 \pm 0.049$&$0.624 \pm 0.043$\\
&\cellcolor{lightgreen} \genemask{} 10K &\cellcolor{lightgreen} \textbf{0.644 $\pm$ 0.008}*&\cellcolor{lightgreen} \textbf{0.695 $\pm$ 0.012}*&\cellcolor{lightgreen} \textbf{0.705 $\pm$ 0.023}**&\cellcolor{lightgreen} \textbf{0.783 $\pm$ 0.023}**&\cellcolor{lightgreen} \textbf{0.639 $\pm$ 0.006}&\cellcolor{lightgreen} \textbf{0.695 $\pm$ 0.008} &\cellcolor{lightgreen} \textbf{0.465 $\pm$ 0.021}&\cellcolor{lightgreen} \textbf{0.651 $\pm$ 0.02}$\dagger$\\ 

500& ORI 10K&\textbf{0.693 $\pm$ 0.010}&\textbf{0.754 $\pm$ 0.007}&$0.751 \pm 0.034$&$0.817 \pm 0.037$&$0.629 \pm 0.006$&$0.679 \pm 0.007$&$0.355 \pm 0.027$&$0.629 \pm 0.014$ \\
&\cellcolor{lightgreen} \genemask{} 10K &\cellcolor{lightgreen} $0.69 \pm 0.006$&\cellcolor{lightgreen} $0.748 \pm 0.009$&\cellcolor{lightgreen} \textbf{0.835 $\pm$ 0.008}**&\cellcolor{lightgreen} \textbf{0.907 $\pm$ 0.008}**&\cellcolor{lightgreen} \textbf{0.648 $\pm$ 0.004}**&\cellcolor{lightgreen} \textbf{0.709 $\pm$ 0.04}**&\cellcolor{lightgreen} \textbf{0.443 $\pm$ 0.025}**&\cellcolor{lightgreen}  \textbf{0.644 $\pm$ 0.015}* \\ 

1000& ORI 10K &$0.705 \pm 0.014$&$0.769 \pm 0.013$&$0.808 \pm 0.015$&$0.884 \pm 0.016$&$0.652 \pm 0.008$& $0.709 \pm 0.008$ & $0.504 \pm 0.02$ & $0.689 \pm 0.016$ \\
&\cellcolor{lightgreen} \genemask{} 10K &\cellcolor{lightgreen} \textbf{0.723 $\pm$ 0.005}&\cellcolor{lightgreen} \textbf{0.785 $\pm$ 0.006}&\cellcolor{lightgreen} \textbf{0.855 $\pm$ 0.004}**&\cellcolor{lightgreen} \textbf{0.925 $\pm$ 0.004}**&\cellcolor{lightgreen} \textbf{0.660 $\pm$ 0.004}*&\cellcolor{lightgreen} \textbf{0.727 $\pm$ 0.004}** &\cellcolor{lightgreen} \textbf{0.507 $\pm$ 0.011}&\cellcolor{lightgreen} \textbf{0.701 $\pm$ 0.007}$\dagger$\\ \hline 
    \end{tabular}
    }
    \caption{Performance comparison for gene sequence classification tasks where all values are rounded to 3 decimal places. The best-performing performance value is written in \textbf{bold}, and the second-best-performing value is highlighted in yellow. \textit{ORI} refers to the SOTA model with random masking. Paired t-test was conducted to check whether the performance improvement of \genemask-based models over ORI 10K model is statistically significant or not; ** for $p$-value < 0.01, * for $p$ < 0.05, and $\dagger$ for p-value < 0.1 }
    \label{tab:perf-compare-10K}
\end{table*}
\raggedbottom

\noindent \textbf{Performance comparison among different masking strategies.} In the case of \textit{DNABert}, \genemaskbest{} \ul{mostly (two out of three times) outperforms all the baseline models across the four datasets of Prom-core, Prom-300, Cohn-enh, and Splice-40}, and in the remaining case where it does not achieve the best performance, it is \ul{always the second-best ranked model} (see Table~\ref{tab:perf-compare-10K}, the second best-performing model is highlighted in yellow). 
We observe that the task-specific performance improvement between \genemaskbest{}{} and random masking (ORI 10K model variant) in \textit{DNABert}, in terms of average accuracy over all few-shot settings for each task, is the highest for Splice-40 followed by Prom-300. Since the \genemask{} (w/ GB) pretraining model variant is explicitly targeted for Splice-40, it highlights the effectiveness of such task-specific interventions.

 In the case of \textit{LOGO}, we do not explore different pretraining model variants and instead compare between ORI 10K and \genemask{} 10K model (both of them have exactly the same model hyperparameters). We observe that \genemask{} 10K (base model) \ul{outperforms random masking-based SOTA models (ORI) in all settings across four benchmark tasks} as shown in Table~\ref{tab:perf-percent-fewshot-ori10K}. 
The highest percentage improvement in terms of average accuracy over all few-shot settings is Splice-40 with $8.32\%$, followed by Prom-300, Prom-core, and Cohn-enh at $7.67\%$, $3.72\%$ and $1.95\%$ respectively. However, we observe the \ul{performance improvement of} \genemask{} \ul{to be higher for LOGO as compared to DNABert for all tasks except the Cohn-enh task}. 

\begin{table}
    \centering
    \begin{tabular}{cccccc}
    \hline
        k-shot & 10 & 50 & 100 & 500 & 1000 \\ \hline
        DNABert &  $2.94\%$ & $0.93\%$ & $0.73\%$ & $0.40\%$ & $1.85\%$\\
        LOGO &  $4.92\%$ & $5.87\%$ & $3.90\%$ & $7.74\%$ & $2.85\%$\\\hline
        
    \end{tabular}
    \caption{Percentage improvement in average accuracy over four datasets of \genemask{} 10K over ORI 10K model. }
    \label{tab:perf-percent-fewshot-ori10K}
\end{table}

\textbf{\genemask{} 10K versus ORI 10K model.}
 In the case of DNABert and LOGO, ORI-10K almost always performs much inferior to any variant of \genemask, although trained for the same number of epochs. Table~\ref{tab:perf-percent-fewshot-ori10K} shows the percentage improvement in average accuracy across four datasets of \genemask{} 10K over the ORI 10K model for various $k$-shot setups. It indicates that \genemask{}-based masking strategy is more beneficial for lightweight models like LOGO; heavier models like DNABert might automatically learn a certain amount of span correlations (like PMI) information, thus diminishing the independent impact of \genemask. 
 From now onwards, we will discuss only DNABert-based models and compare the performance of \genemaskbest{} with the baseline models.  

\noindent \textbf{ \genemaskbest{} versus Gene-PMI-VOC model.} 
We observe that \genemaskbest{} \ul{almost consistently outperforms the Gene-PMI-VOC model, except for the 10, 50, and 100-shot settings of the Splice-40 task}. We further observe that the performance of \ul{Gene-PMI-VOC is quite unstable as it even performs poorer than the ORI 10K model in most cases} (12 out of 20 DNABert settings in Table~\ref{tab:perf-compare-10K}). A possible reason for such inconsistent performance is its sensitivity to the (arbitrary) size of its masking vocabulary, created using a PMI score to formulate an absolute importance notion. However, determining the optimal vocabulary size in a principled manner is difficult due to the absence of human-understandable semantics in genomics (like words and phrases in NLP). A suboptimal choice may lead to difficulty in guaranteeing diverse masking patterns, thus ushering in an inefficient MLM training regime. 

\begin{table}
    \centering
    \begin{tabular}{cccccc}
    \hline
        k-shot & 10 & 50 & 100 & 500 & 1000 \\ \hline
        ORI 120K & $3.58\%$ & $2.46\%$ & $2.64\%$ & $10.17\%$ & $8.15\%$ \\ \hline
    \end{tabular}
    \caption{Percentage improvement in average accuracy over four datasets of \genemaskbest{} over ORI 120K model in case of DNABert.}
    \label{tab:perf-percent-fewshot-ori120K}
\end{table}

\noindent \textbf{ \genemaskbest{} versus ORI-120K model.} Table~\ref{tab:perf-percent-fewshot-ori120K} shows the percentage improvement in average accuracy scores (over four datasets) of \genemaskbest{} over the original DNABert model trained for 120K steps (ORI 120K model). \genemaskbest{} improves decently over ORI 120K in low data settings (100-shot and below) but the performance rises in 500 and 1000-shot settings, which is contrary to our standard observation (all datasets except Splice-40). This happens due to the Splice-40 task, where the performance improvement due to \genemaskbest{} increases at higher data settings ($0.692$ versus $0.429$ and $0.709$ versus $0.496$ for \genemaskbest{} and ORI 120K model respectively) as \genemask{} (w/ GB) is effective at that setting. 
Another interesting observation is that \ul{ORI 120K model outperforms} \genemaskbest{} \ul{on shallow data settings (10, 50, and 100-shot) of the Prom-core task, whereas} \genemaskbest{} \ul{outperforms again in 500 and 1000-shot settings}. This may be because Prom-core is an easier task with a much shorter context (70 base pairs (bp) in length as compared to 300 bp for Prom-300 and 500 bp for Cohn-enh), whereby the ORI 120K model simply memorizes the gene sequence patterns instead of actually learning intrinsic (or extra) knowledge of gene sequences. Hence, the effect of memorization recedes in higher data settings (above 500-shot), as seen in Figure~\ref{fig:perf-compare-graph} of Appendix. The reason behind the improvement in the Splice-40 task at a higher $k$-shot is also similar. 

\noindent \textbf{Deconstructing \genemaskbest{}.}  We compare the performance of three model variants of \genemask{}. We observe from Table~\ref{tab:ablation-compressed} that \ul{half gradient accumulation (HGA) helps to improve the model performance in 10 and 50-shot} settings (10 and 50-shot for Prom-300, 50-shot for Cohn-enh, and 10-shot for Splice-40). It is empirically observed that if large batch size is used to train deep neural networks, the trained models appear to generalize poorly~\cite{hoffer2017,keskar2017}. HGA reduces the effective batch size by $50\%$ and mitigates the \textit{generalization gap} issue in such low data settings (10 and 50-shot). However, beyond the 100-shot setting, the impact is negative or marginal. We observe that when we consider \ul{gene boundaries during pretraining data construction (GB)}, the performance consistently improves for the Splice-40 task. The primary reason may be that the \ul{Splice-40 task involves identifying exon boundaries, which can benefit strongly from gene boundaries as side information}.  

\begin{table}[!ht]
    \footnotesize
    \centering
    \scalebox{0.72}{
    \addtolength{\tabcolsep}{-0.3em}
    \begin{tabular}{cccccc}
    \hline
        \textbf{k-shot} &  \textbf{Model Type}  & \textbf{Prom-core}& \textbf{Prom-300} & \textbf{Cohn-enh} & \textbf{Splice-40}  \\ \hline

10 & \cellcolor{lightgreen} \genemask&\cellcolor{lightgreen} \textbf{0.602 $\pm$ 0.058}&\cellcolor{lightgreen} $0.625 \pm 0.09$&\cellcolor{lightgreen} \textbf{0.622 $\pm$ 0.050}&\cellcolor{lightgreen} $0.392 \pm 0.02$\\

& \genemask w/ HGA &$0.585 \pm 0.065$&\textbf{0.676 $\pm$ 0.054}&$0.604 \pm 0.068$&\textbf{0.412 $\pm$ 0.026} \\

& \genemask w/ GB&$0.568 \pm 0.05$ &$0.593 \pm 0.038$&$0.602 \pm 0.043$&$0.398 \pm 0.015$ \\ 

50 & \cellcolor{lightgreen} \genemask &\cellcolor{lightgreen} \textbf{0.678 $\pm$ 0.026}&\cellcolor{lightgreen} $0.781 \pm 0.097$&\cellcolor{lightgreen} $0.648 \pm 0.016$&\cellcolor{lightgreen} $0.505 \pm 0.018$  \\ 

& \genemask w/ HGA &$0.677 \pm 0.033$&\textbf{0.815 $\pm$ 0.02}&\textbf{0.654 $\pm$ 0.017}&$0.448 \pm 0.095$ \\

&  \genemask w/ GB &$0.669 \pm 0.025$&$0.793 \pm 0.022$&$0.641 \pm 0.012$&\textbf{0.519 $\pm$ 0.027} \\ 

100 & \cellcolor{lightgreen} \genemask&\cellcolor{lightgreen} \textbf{0.708 $\pm$ 0.013}&\cellcolor{lightgreen} $0.843 \pm 0.027$&\cellcolor{lightgreen} $0.655 \pm 0.052$&\cellcolor{lightgreen} $0.561 \pm 0.027$ \\ 

& \genemask w/ HGA &$0.694 \pm 0.043$&\textbf{0.847 $\pm$ 0.029}&\textbf{0.67 $\pm$ 0.017} &$0.555 \pm 0.079$ \\
& \genemask w/ GB&$0.695 \pm 0.014$&$0.829 \pm 0.016$&$0.661 \pm 0.008$&\textbf{0.577 $\pm$ 0.019} \\ 

500 & \cellcolor{lightgreen} \genemask &\cellcolor{lightgreen} \textbf{0.753 $\pm$ 0.005}&\cellcolor{lightgreen} \textbf{0.89 $\pm$ 0.006}&\cellcolor{lightgreen} \textbf{0.698 $\pm$ 0.006} &\cellcolor{lightgreen} $0.669 \pm 0.017$ \\  

& \genemask w/ HGA &$0.75 \pm 0.008$&$0.887 \pm 0.004$&$0.674 \pm 0.058$ &$0.662 \pm 0.026$ \\

&  \genemask w/ GB&$0.746 \pm 0.006$&$0.877 \pm 0.005$&$0.675 \pm 0.005$&\textbf{0.692 $\pm$ 0.016} \\ 

1000&  \cellcolor{lightgreen} \genemask &\cellcolor{lightgreen} \textbf{0.766 $\pm$ 0.004}&\cellcolor{lightgreen} \textbf{0.898 $\pm$ 0.005}&\cellcolor{lightgreen} \textbf{0.706 $\pm$ 0.005}&\cellcolor{lightgreen} \textbf{0.709 $\pm$ 0.012} \\ 

& \genemask w/ HGA &\textbf{0.766 $\pm$ 0.007}&\textbf{0.898 $\pm$ 0.004}&$0.699 \pm 0.008$ &$0.69 \pm 0.025$ \\

& \genemask w/ GB&$0.756 \pm 0.005$&$0.889 \pm 0.005$&$0.688 \pm 0.004$&$0.693 \pm 0.02$ \\ \hline 
    \end{tabular}
    }
    \caption{Performance comparison in terms of accuracy among \genemask-guided DNABert model variants at $10000$ steps. \textit{\genemask} represents the hyperparameter settings of baseline models as ORI 10K model}
    \label{tab:ablation-compressed}
\end{table}
\raggedbottom

\section{Domain-specific Model Explainability}
\label{sec:motifs}
\begin{table}
    \centering
    \scalebox{0.9}{
    \begin{tabular}{cccc} 
    \hline
         Dataset&Motifs& Normalized PMI rank  \\
         &(Consensus Logo)& (out of 4096) \\ \hline
         Prom-core & n\textbf{TATAAA}r &242\\
         Cohn-enh & \textbf{GTGGCT}sw & 126 \\
         Prom-core, Cohn-enh&nCyy\textbf{CCTCC}n* &1, 11, 52, 175, 186 \\
         Prom-core & sCw\textbf{GCAGC}n & 259, 516, 540, 570, 628 \\
         Cohn-enh & ks\textbf{CTGGG}m &5, 17, 20, 21, 71 \\
         Cohn-enh & \textbf{TTTTTT}TTTn & 8 \\ \hline
    \end{tabular}
    }
    \caption{PMI-based rankings based on Normalized-PMI\textsubscript{n} score for motifs in finetuning datasets. A motif of length five matches as a sub-string to multiple 6-mers and thus mention the top five ranks}
    \label{tab:motif-6-mer-ranking}
\end{table}

Motifs are repetitive units within a \textit{Human Reference Genome}, having a certain biological significance. We check whether the correlated tokens identified by \genemask{}  are, in fact (part of) such units. We also perform a  preliminary evaluation on the task of identifying functional genetic variants in Section~\ref{sec:app-func-gene-variants}.

\noindent \textbf{Associating top-20 ranked 6-mers based on Normalized-PMI\textsubscript{k} score with conserved DNA sequence motifs.} 
We analyze whether \ul{highly ranked PMI tokens} resemble meaningful concepts by checking their \ul{overlap with known motifs}. We performed a Google search with the following query template: [``6-mer name'' \textit{DNA sequence motif}]. \textit{AATCTC} is a 6-mer, ``'' are used as a Google wildcard to indicate that the term \textit{AATCTC} must always be present in search results. We only considered the first page of Google results to determine whether a particular 6-mer is mentioned in biomedical literature.  Among the \textit{top 20 ranked} 6-mers, we observed that all except 2 (\textit{CCAGGC} - rank nine, \textit{GCCTGG} - rank ten) are indeed previously mentioned in the biomedical literature. Since the \genemask{} strategy favors the top-ranked PMI tokens that correlate well with the known DNA sequence motifs of enhancers and promoters, this knowledge translates to improved performance in low data settings (10, 50, and 100-shot) on the Prom-core, Prom-300, and Cohn-enh datasets. 

\noindent \textbf{PMI-based rankings capture motifs present in finetuning datasets.} We use the R package \textit{rGADEM}~\cite{rgadem} to perform de novo motif discovery and obtain a total of $12$ motifs from the Prom-core and Cohn-enh datasets, which are mostly concentrated around lengths 5, 6, and 7. We present our best matches and their corresponding ranks in the 6-mer PMI ranked list (\textit{RANK}) in Table~\ref{tab:motif-6-mer-ranking} (see Table~\ref{tab:appendix-motif-6-mer-ranking} for the complete list, their corresponding consensus logos in Figures~\ref{fig:consensus-logos-1} and~\ref{fig:consensus-logos-2}). 
We observe that most of the \ul{6-mers that match the discovered motifs are ranked very high}. We further observe that our \ul{top-1 ranked 6-mer} is present in motifs of both \ul{enhancers and promoters} based on de-novo motif discovery and is mentioned in previous biomedical literature~\cite{chow1991}. The TATA box, a well-known motif for promoters (row 1 of Table~\ref{tab:motif-6-mer-ranking}) is ranked at $242$; the best TATA box motif is \textit{TATATA} with a PMI rank of $15$. 
\section{Conclusion}
In this paper, we develop a PMI- induced novel masking algorithm, \genemask{}, for MLM training of gene sequences which ensures a substantial speedup of 10x and performance improvement over the random masking strategy of SOTA models (DNABert and LOGO) in various few-shot settings. 
Further we showed that for a certain task (Splice-40), respecting gene boundaries during pretraining data construction improves the performance even further, confirming the basic thesis of the importance of incorporating domain-specific information during pretraining. This is further corroborated when we observe a strong correlation  between top-ranked NPMI tokens and conserved DNA sequence motifs, implying that \genemask{} is able to capture meaningful semantic structure in gene sequences. Finally, we must mention that gene sequence classification is a challenging problem and requires a tremendous engineering effort even to develop the experimental setup. Therefore, the codes (including trained models), datasets, and appendix are made publicly available at \url{https://github.com/roysoumya/GeneMask}. 
The elaborate setup developed will help us undertake several 
 \textbf{future work}, for example, we will explore time-variant MLM training strategies for gene sequences to achieve better few-shot performance, such as incorporating a decaying masking ratio, which showed good performance on GLUE and SQuaD datasets in NLP domain~\cite{yang2023learning}. 
 
 \ack Soumyadeep Roy is supported by the Institute Ph.D. Fellowship at the Indian Institute of Technology Kharagpur. Soumyadeep Roy and Niloy Ganguly were also affiliated with L3S Research Center, Germany while conducting this work. This research was funded by the Federal Ministry of Education and Research (BMBF), Germany, under the project LeibnizKILabor with grant No. 01DD20003.

\bibliography{ecai}
\newpage
\newpage
\appendix
\section{Appendix}
In this section, we provide the supplementary material associated with the paper. 

\subsection{Ethics Statement}
The gene sequence data used for pretraining and finetuning is obtained from publicly available sources and can be obtained directly without signing any explicit data use agreement. The three benchmark datasets of Prom-core, Prom-300, and Cohn-enh are also used in previous studies for the task of gene sequence classification~\cite{bigbird2020,DNABert,martinek2022,deepromoter2019}. Our work does not involve patient-level data for the experiments.  We do not foresee any negative social impacts of this work, but of course, the accumulation of improvements in ML could be misused as it may give more power to nefarious agents.

\subsection{Background}
\noindent \textbf{Importance of understanding gene regulatory code.}
The long strands of DNA found in the human chromosomes can be classified into genes, and the genes, in turn, comprise \textit{coding} and \textit{non-coding} parts. A \textit{coding} part encapsulates the information required for converting the nucleotide to a \textit{protein}. These proteins are the building blocks of all tissues. These genes interact with the non-coding regions which perform gene regulation. \textit{Promoters/Enhancers} speed up the process of coding, \textit{inhibitors} slow down the reaction. These non-coding genes are called \textit{gene regulatory elements}\cite{genomics-survey}. The non-coding regions, accounting for over $98\%$ of the whole genome, implement significant yet largely unknown regulatory functions. Recent large consortia projects, including the ENCyclopedia of DNA Elements (ENCODE)~\cite{encode2012integrated}, Roadmap Epigenomics~\cite{kundaje2015integrative}, and the Genomics of Gene Regulation (GGR), have produced a large number of experimental mapping readouts to help annotate non-coding genome in specific tissues or cell-lines. On the other hand, Genome-wide association studies (GWAS) have discovered that the vast majority ($>90\%$) of associated genome loci for complex diseases and traits fall in non-coding regions~\cite{visscher2017}.

\noindent \textbf{Learning deep representations in the context of gene sequence modeling.}
 Nguyen et al.~\cite{nguyen2016dna} encode base pair triples as one-hot vectors to feed into convolutional neural networks for DNA sequence classification tasks, whereas Badirli et al.~\cite{Badirli2021} convert the DNA barcodes represented by nucleotide sequences into a vector embedding useful for the task of fine-grained species classification.

\noindent \textbf{MLM training.}
\cite{yamaguchi-etal-2021-frustratingly} explore alternative pretraining tasks compared to MLM, such as shuffled word detection, random word detection, manipulated word detection (Shuffle + Random), masked token type classification, and masked first character prediction. Here, we choose the original DNABert configuration of MLM without Next Sentence Prediction and experiment with multiple masked token selection strategies. 

\subsection{Methodology}

\subsubsection{Gene-PMI-VOC model: Limitations of Trivial Adaptation of PMI Masking to Gene Sequence}\label{sec:genepmivoc-details}

We initially developed the closest adaptation of the standard PMI masking~\cite{pmi-masking} approach used in NLP to the gene sequence classification setup, which involved significant human effort from our end. We consider it as a baseline model (Gene-PMI-VOC) for our experiments. We describe the working of the model in detail in Algorithm~\ref{algo:genepmivoc}. We next describe the various drawbacks of the \textit{Gene-PMI-VOC} model: 

\begin{itemize}
    \item  \textbf{Determining the masking vocabulary size:} Levine et al.~\cite{pmi-masking} use a small-scale evaluation of an n-gram's collocation quality as a function of its rank. They create an ad-hoc dataset composed of 1000 n-grams that they manually label as the collocated terms or not (for more details, please see Section A of the Appendix of~\cite{pmi-masking}). Creating such an ad-hoc dataset in the gene sequence modeling setting is difficult because: (i) Gene sequences are continuous, and such semantically-meaning unit (like words or tokens in NLP) is not available in the genomic domain. (ii) The manual annotation process will require medical domain experts' involvement and, thus, a significant human effort.

   \item The standard PMI formulation~\cite{pmi-masking} \textbf{favors tokens with a low frequency of occurrence}. This significantly reduces the efficacy of PMI masking and leads to the wastage of PMI masking-based pretraining steps because, during the MLM training stage, such high PMI tokens will rarely appear as masked tokens (given their low frequency of occurrence). 
    
\end{itemize}

\SetCommentSty{mycommfont}
\SetKwInput{KwInput}{Input}      
\SetKwInput{KwOutput}{Output}    
\SetKwInput{KwData}{Initialization}

\LinesNumberedHidden

\begin{algorithm}[h]
\footnotesize
\caption{Gene-PMI-VOC Algorithm.}\label{algo:genepmivoc}

\DontPrintSemicolon \;
    \KwInput{DNA string made of 6-mer tokens with a maximum length of 510, $MaskVocab$: Fixed Masking Vocabulary composed of k-mers where $1 \leq k \leq 10$, $m$: MLM probability = $0.025$ }
    \KwOutput{$MaskTokenSet$: List of tokens to mask in the DNA string}
    \KwData{
    
    \tcp{P[i] denotes the PMI-token at $i$-th position in the DNA segment, DNA[i] denotes the $i$-th nucleotide, $p =$ maximum input sequence length * MLM probability $= 512 * 0.025 = 12.8$ on average, $MaskVocab$ size $= 40000$, $PmiTokenizedString$: stores the DNA input sequence tokenized by PMI tokenizer}

    $MaskTokenSet \gets \emptyset$

    $PmiTokenizedString$ $\gets \emptyset$
  
    }
	\SetKwFunction{FMain}{MapNucleotideToKmerTokens}
	\SetKwProg{Fn}{Function}{:}{}
	
	\Fn{\FMain{nucleotide position id $i$}}{
            
            $MappedTokens$ $\gets$ T[$j$],  $\forall(j)_{j=i - 2}^{i + 3}$ \;
			
			\KwRet\  $MappedTokens$ \; 
    }

    \textbf{Step 1:} Tokenize the input DNA string into PMI tokens obtained from the fixed vocabulary $MaskVocab$. Select the 10-mer from the start of the input DNA string $s$ and check whether it is in $MaskVocab$. If found true, add it to $PmiTokenizedString$, move to the end of the 10-mer, and repeat the process until the end of $s$ is reached. If found false, repeat the same thing with 9-mer, 8-mer, ..., and 1-mer consecutively until there is a positive match with $MaskVocab$.

    \textbf{Step 2:} Randomly select $p$ nucleotides as mask centers (MC) spread uniformly over the DNA string.
            
    \textbf{Step 3:}
    \For{each nucleotide in $p$ mask centers}
	{
            $PositionId$ := Token index of nucleotide on the DNA segment given as input
            
            $PmiNucleotides$ := Using $PmiTokenizedString$, identify the PMI gene sequence corresponding to the nucleotide at $PositionId$)

            \For{each nucleotide $d$ in $PmiNucleotides$}
            {
                $MaskTokenSet$ $\gets$ $MaskTokenSet$~$\cup$~ $MapNucleotideToKmerTokens$ ($d$)
                }
                     
	} 

     \KwRet\  $MaskTokenSet$ \;
\end{algorithm}

\begin{table}
\centering
\small
\begin{tabular}{cc}
\hline
Sequence Length  & Total tokens in vocabulary  \\ \hline
1  & 5  \\
2  & 17  \\
3  & 65  \\
4 & 212  \\
5  & 533  \\
6 & 1465    \\
7   & 3829   \\
8   & 10271   \\
9& 17537   \\
10    & 6071  \\ \hline
\end{tabular}
\caption{Masking vocabulary statistics of Gene-PMI-VOC baseline model}
\label{tab:pmi-vocab-dist}
\end{table}

\subsubsection{Normalized PMI\textsubscript{k} Metric Details.} Here, we provide more implementation details for the Normalized \textit{PMI\textsubscript{k}} (N-PMI) as defined in Equation~\ref{eq:norm-pmi}. We choose the minimum frequency of occurrence threshold $c$ as $101$, which puts a cut-off beyond the first quartile ($25$ percentile) of k-mer frequencies. The k-mer frequency distribution follows a long-tailed distribution; hence the number of tokens below 101 is very low. 

\subsection{Experimental Setup: Training Details}\label{sec:appendex-model-parameter}

\noindent \textbf{Finetuning parameter configuration.} The models are finetuned for $20$ epochs at a learning rate of $4e^{-4}$, warmup steps percentage of $10\%$, hidden dropout probability of $0.1$, weight decay as $0.01$, per GPU train batch size as $5$ and use the \textit{AdamW} optimizer for the 10, 50 and 100-shot setting. However, the performance drops in 500 and 1000-shot settings due to overfitting. Therefore, for the 500-shot settings and above, we use the same hyperparameters as the original DNABert paper --- a lower learning rate from $4e^{-4}$ to $5e^{-5}$ and a lower number of epochs from 20 to 5. Mosbach et al.~\cite{mosbach2021} observe that such a high number of fine-tuning epochs helps address random initialization issues in the low-resource settings (10, 50, and 100-shot). 

\begin{table}[!ht]
    \centering
    \begin{tabular}{ccc}
    \hline
        \textbf{Parameter} & \textbf{DNABert} & \textbf{LOGO} \\
        \hline
        Hidden Size & 768 & 256 \\
        Hidden Layers & 12 & 2 \\
        Attention Heads & 12 & 8 \\
       Per GPU train batch size & 10 & 5 \\
        Hidden Dropout Probability & 0.1 & 0 \\
        Attention Dropout Probability & 0.1 & 0 \\
        Intermediate Size & 3072 &  3072 \\
        Embedding Size & 512 & 512 \\ \hline
    \end{tabular}
    \caption{Difference between parameters of DNABert and LOGO}
    \label{tab:dna-logo-config}
\end{table}

\subsection{Experimental Results: Memorization versus Generalization}\label{sec:expt-results}

\begin{figure*}
    \centering
    \includegraphics[width=\textwidth]{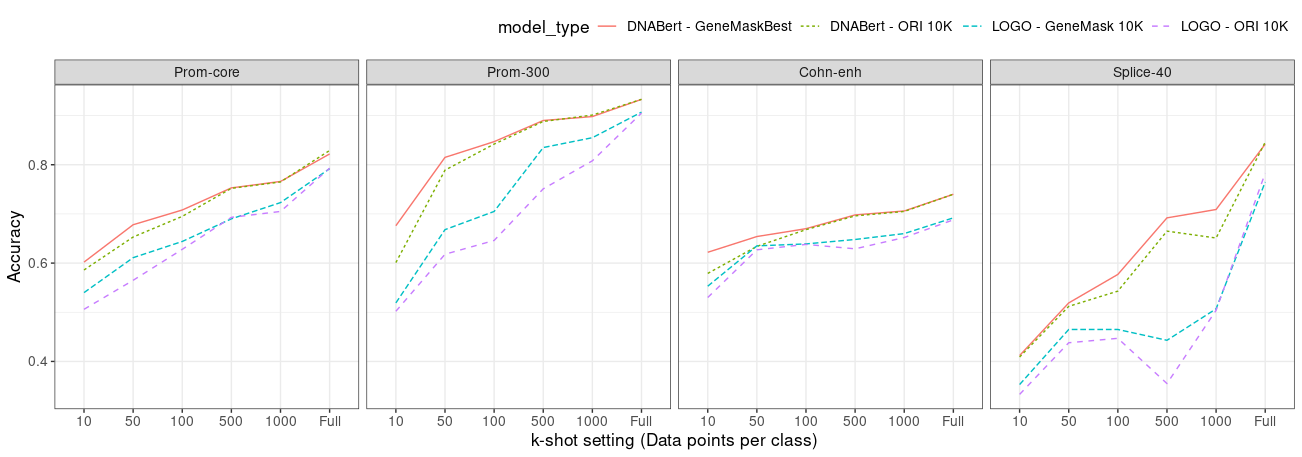}
    \caption{Model performance comparison across different few-shot settings and full data over four benchmark datasets}
    \label{fig:perf-compare-graph}
\end{figure*}

It was observed in Radford et al.~\cite{radford2019language} that data overlap between WebText training data and specific evaluation datasets like Winograd Schema Challenge, CoQA, and LAMBADA, provides a small but consistent improvement to model performance; this indicates that the model generalization results suffer from severe over-reporting. Instead of using a probabilistic data structure like Bloom filters as used by Radford et al.~\cite{radford2019language}, we use a deterministic method based on sets and hashing to compute the exact overlap between the pretraining data with task-specific test data and between task-specific train and task data. Specifically, we compute the percentage of $15$-grams (a subsequence of k-mers of length $15$) that are common between the dataset pairs. This leads to $185.4$ million unique $15$-grams for the pretraining data (based on the Human Reference Genome), which covers $17.26\%$ of all possible $15$-length combinations. The overlap results are presented in Table~\ref{tab:data-overlap}. 

\ul{We observe that the task-specific train and test data overlap is very low (median overlap of $\leq 5.2\%$) for all the datasets except for Prom-300 (median overlap of $31.9\%$).} To investigate the contribution to model performance due to memorization, we remove test data points with more than $50\%$ overlap or top five percentile data points in terms of the highest overlap ratio. We observe that for all tasks except Splice-40, there is a marginal drop in performance 
whereas, for Splice-40, performance improves marginally (please see the first and second rows of Table~\ref{tab:perf-compare-10K}). \ul{We thus conclude that although there is some overlap between the task-specific train and test dataset, it does not contribute much to model performance.} Thus, memorization in DNABert architecture does not lead to over-reporting of generalization performance for the models discussed in this paper. Furthermore, we evaluate only in few-shot settings, where only a small portion of the training dataset 
is used for finetuning.

\begin{table}
    \centering
    \small
    \begin{tabular}{ccc} \hline
        Task&Training data & Pretraining data \\ \hline
         Prom-core&0.018 (0.255)& 0.218 (0.655) \\ 
         Prom-300&0.319 (0.621) & 0.274 (0.621) \\ 
         Splice-40& 0.0 (0.24) & 0.052 (0.723) \\
         Cohn-enh & 0.569 (0.956)& 0.583 (0.953)\\ \hline
        
    \end{tabular}
    \caption{Overlap ratio of test dataset with pretraining data and training data. The median is reported along with the $95$ percentile value in brackets}
    \label{tab:data-overlap}
\end{table}

\subsection{Domain-specific Model Explainability: PMI-based rankings capture motifs present in fine-tuning datasets}
\begin{figure}
    \centering
    \includegraphics[width=0.45\textwidth, height=3cm]{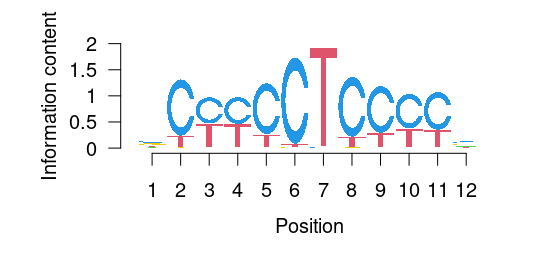} 
    \includegraphics[width=0.45\textwidth, height=3cm]{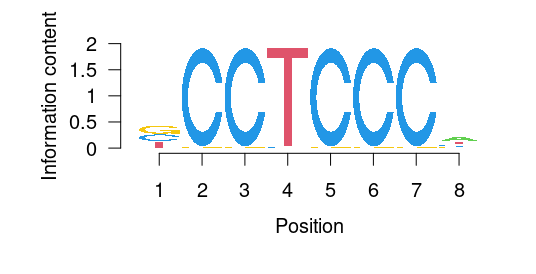} 
    \includegraphics[width=0.45\textwidth, height=3cm]{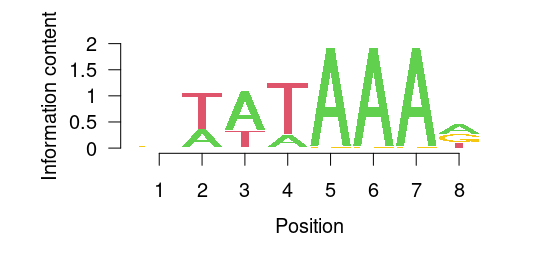} 
    \caption{Consensus logo plot of motifs identified using de novo motif discovery tool. (left) nCyyCCTCCyCn (middle) sCCTCCCw (right) nTATAAAr }
    \label{fig:consensus-logos-1}
\end{figure}

\begin{figure}
    \centering
    \includegraphics[width=0.45\textwidth, height=3cm]{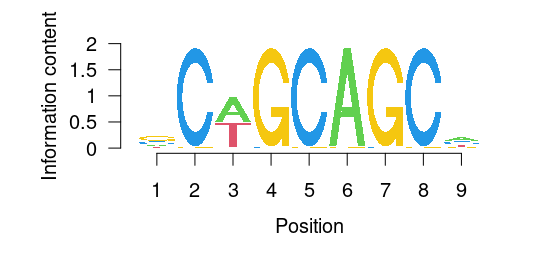}  
    \includegraphics[width=0.45\textwidth, height=3cm]{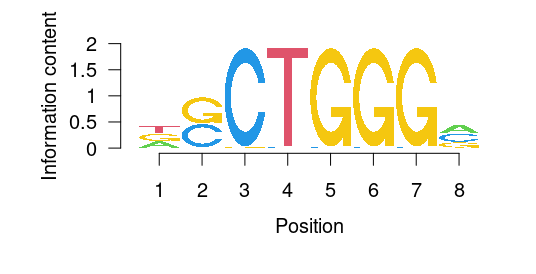} 
    \includegraphics[width=0.45\textwidth, height=3cm]{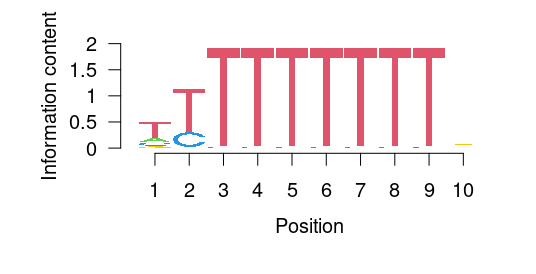}  
    \caption{Consensus logo plot of motifs identified using de novo motif discovery tool. (left) sCwGCAGCm (middle) ksCTGGGm (right) TTTTTTTTTn}
    \label{fig:consensus-logos-2}
\end{figure}

As motif discovery is computationally expensive, we only provide a subset of the data: randomly sampled 1000 (prom300 - $300$ bp and enhancers-cohn $500$ bp) and $2000$ (prom-core - 70 bp) data points. We present the complete list of matches and their corresponding ranks in the 6-mer PMI ranked list (\textit{RANK}) in Table~\ref{tab:appendix-motif-6-mer-ranking}, their corresponding consensus logos in Figures~\ref{fig:consensus-logos-1} and~\ref{fig:consensus-logos-2}.

\begin{table}
    \centering
    \scalebox{0.7}{
    \begin{tabular}{ccc}\hline
         Dataset&Motifs& Normalized PMI rank \\ && (Top 5 for 5-mers) (out of 4096) \\ \hline
         Prom-core&nCyy\textbf{CCTCC}n* &1, 11, 52, 175, 186  \\
         Prom-core&sCs\textbf{CCGCC}sCCn & 103, 1181, 1678, 2205, 2534 \\
         Prom-core&sCw\textbf{GCAGC}n & 259, 516, 540, 570, 628  \\
         Prom-core&yy\textbf{TTTATA}n & 286  \\
         Prom-core&n\textbf{TATAAA}r &242 \\
         Prom-core&n\textbf{GAGGAGG}v & AGGAGG (rank 56), GAGGAG (rank 278) \\
         Prom-core&k\textbf{GCTGC}wGs & 260, 510, 555, 590, 639 \\
         Cohn-enh&ks\textbf{CTGGG}m &5, 17, 20, 21, 71 \\
         Cohn-enh&n\textbf{CCTGGCC}h & CCTGGC (rank 25), CTGGCC (rank 129)  \\
         Cohn-enh&yy\textbf{CCAG}r\textbf{G}n & 302, 593, 1247, 2778 \\
         Cohn-enh&\textbf{TTTTTT}TTTn & 8 \\
         Cohn-enh&\textbf{GTGGCT}sw & 126 \\
         \hline
    \end{tabular}}
    \caption{PMI-based rankings based on NPMI\textsubscript{k} 
    score for the motifs present in finetuning datasets. The motifs are of lengths 5, 6, or 7. For length 7, we mention two rankings considering two 6-length sub-motifs. A motif of length five matches as a sub-string to multiple 6-mers; we only mention the top five ranks for all such matches.}
    \label{tab:appendix-motif-6-mer-ranking}
\end{table}

\subsubsection{Analyzing the effect of functional genetic variants}\label{sec:app-func-gene-variants} 
Here, we aim to test the extent of domain-specific knowledge learned by the \genemask-guided models over the original SOTA model (ORI). Therefore, instead of the gene sequence classification task, we investigate the model performance for the task of identifying functional genetic variants. We thus reproduce the variant analysis conducted by~\cite{DNABert}, using dbSNP~\cite{dbsnp2001} and ClinVar~\cite{clinvar2013}, to compare the performance of the \genemask{} model with the original SOTA model (ORI), when both models are finetuned on 10-shot Prom-core dataset. 

 $400,000$ variants were retrieved from dbSNP, and the original authors constructed the corresponding genomic sequences (both original and mutated). When the original and mutated sequences offer significantly different prediction probabilities with respect to promoter prediction, the variant is queried in ClinVar and other similar databases to ascertain their importance. Analyzing these queries established that the variants identified by the fine-tuned promoter model have \ul{interpretable uses and links to diseases or other functional aspects}. As a result, DNABERT demonstrated its capability to \ul{capture and propose new and significant (disease-specific) variants in the future.} 

\noindent \textbf{Implementation Details.} Since we do not have access to the specific code used by the authors to evaluate the importance of a given variant using Clinvar, we instead compare our model with the ORI 120K model variant of DNABert, which is also fine-tuned on the 10-shot Prom-core dataset (ORI 120K performs best in Prom-core low-resource setting) and obtain the original weights used by the authors~\cite{DNABert-Codebase}.  The differences in promoter prediction probabilities for the dataset mentioned above ($400,000$ data points) are recorded. Then, the sequences are ranked by non-increasing order of the absolute difference value. Using the published values given by the authors for ORI 120K promoter setting as ground truth (i.e., a proxy to identify all possible important functional variants), we compare a model's ranked list of variants. At this point, we obtain an individual ranked list of data points for the ORI and PMI \textit{few-shot} models. 

\begin{figure}[t]
\centering
\includegraphics[width=0.4\textwidth]{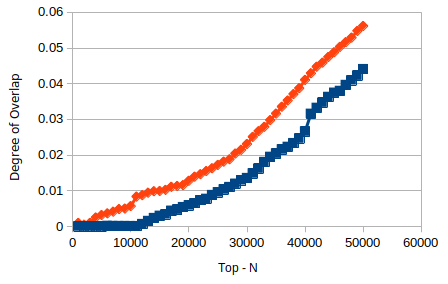}
\caption{Performance comparison between ORI 10K (blue line) and \genemask{} 10K model (orange line) in terms of overlap percentage with the ground-truth ranked list for the task of analyzing the effect of functional genetic variants}
\label{fig:variant-perf-comp}
\end{figure}

\noindent \textbf{Discussion of Results.} The degree of overlap (intersection of the two lists) with the ground-truth ranked list for multiple top-N settings for the ORI 10K and \genemask{} 10K setting is provided in Figure~\ref{fig:variant-perf-comp}. We observe that \genemask-guided \ul{DNABert consistently reports higher overlap over the original model (ORI) in different top-N settings ($5000 \leq N \leq 50000$)}. Thus, we conclude that \genemask{} helps incorporate intrinsic (or relevant) genomics information into DNABert. 

\if{0}
\section{Page limit}
%

The page limit for ECAI scientific papers is {\bf 7} pages, plus one ({\bf 1})
additional page for references only. Scientific papers should report on substantial novel results. The reference list may start earlier than page 8, but only references are allowed on this additional eighth page.
The page limit for ECAI highlights is {\bf 2} pages. They are intended for disseminating  recent technical work (published elsewhere), position, or open problems with clear and concise formulations of current challenges.

Please consult the most recent Call For Papers (CFP) for the most up-to-date
detailed instructions.

Page limits are \textit{strict}. Overlength submissions will be rejected without review.

\section{General specifications}
The following details should allow contributors to set up the general
page description for their paper:

\begin{enumerate}
\item The paper is set in two columns each 20.5 picas (86 mm) wide
  with a column separator of 1.5 picas (6 mm).

\item The typeface is Times Modern Roman.

\item The body text size is 9 point (3.15 mm) on a body of 11
point (3.85 mm) (i.e., 61 lines of text).

\item The effective text height for each page is 56 picas (237 mm).
The first page has less text height. It requires an additional footer
space of 3.5 picas (14.8 mm) for the copyright inserted by the publisher
and 1.5 picas (6 mm) of space before the title.
The effective text height of the first page is 51 picas (216 mm).

\item There are no running feet for the final camera-ready version of the
paper. The submission paper should have page numbers in the running feet.

\end {enumerate}

\section{Title, author, affiliation, copyright
and running feet}
\subsection{Title}
The title is set in 20 point (7 mm) bold with leading of 22 point (7.7
mm), centered over the full text measure, with 1.5 picas (6 mm) of
space before and after.

\subsection{Author}
The author's name is set in 11 point (3.85 mm) bold with leading of 12
point (4.2 mm), centered over full text measure, with 1.5 picas (6 mm)
of space below. A footnote indicator is set in 11 point (3.85 mm)
medium and positioned as a superscript character.

\subsection{Affiliation}
The affiliation is set as a footnote to the first column. This is set
in 8 point (2.8 mm) medium with leading of 8.6 point (3.1 mm), with a
1 point (0.35 mm) footnote rule to column width.

\subsection{Copyright}
The copyright details will be inserted by the publisher.

\subsection{Running feet}
The running feet are inserted by the publisher. For submission you may
insert page numbers in the middle of the running feet. Do not,
however, insert page numbers for the camera-ready version of the
paper.

\section{Abstract}
The abstract for the paper is set in 9 point (3.15 mm) medium, on a
body of 10 point (3.5 mm). The word Abstract is set in bold, followed
by a full point and a 0.5 pica space.

\section{Headings}\label{heads}
Three heading levels have been specified:

\begin{enumerate}
\item{A level headings}

\begin{itemize}
\item The first level of heading is set is 11 point (3.85 mm) bold, on
  a body of 12 point (4.2 mm), 1.5 lines of space above and 0.5 lines
  of space below.

\item The heading is numbered to one digit with a 1 pica space
  separating it from the text.

\item The text is keyed in capitals and is unjustified.
\end{itemize}

\item{B level headings}
\begin{itemize}
\item The second level of heading is set is 11 point (3.85 mm) bold,
  on a body of 12 point (4.2 mm), 1.5 lines of space above and 0.5
  lines of space below.

\item The heading is numbered to two digits separated with a full
  point, with a 1 pica space separating it from the text.

\item The text is keyed in upper and lower case with an initial
  capital for first word only, and is unjustified.
\end{itemize}

\item{C level headings}
\begin{itemize}
\item The third level of heading is set is 10 point (3.5 mm) italic,
  on a body of 11 point (3.85 mm), 1.5 lines of space above and 0.5
  lines of space below.

\item The heading is numbered to three digits separated with a full
  point, with a 1 pica space separating it from the text.

\item The text is keyed in upper and lower case with an initial
  capital for first word only, and is unjustified.
\end{itemize}

\item{Acknowledgements}

  This heading is the same style as an A level heading but is not
  numbered.
\end{enumerate}

\section{Text}
The first paragraph of text following any heading is set to the
complete measure (i.e., do not indent the first line).

Subsequent paragraphs are set with the first line indented
by 1 pica (3.85 mm).

There isn't any inter-paragraph spacing.

\section{Lists}
The list identifier may be an arabic number, a bullet, an em
rule or a roman numeral.

The items in a list are set in text size and indented by 1
pica (4.2 mm) from the left margin. Half a line of space is
set above and below the list to separate it from surrounding
text.

See layout of Section \ref{heads} on headings to see the results of the list macros.

\section{Tables}
Tables are set in 8 point (2.8 mm) on a body of 10 point (3.5 mm).
The table caption is set centered at the start of the table, with
the word Table and the number in bold. The caption is set in medium
with a 1 pica (4.2 mm) space separating it from the table number.

A one line space separates the table from surrounding text.

\begin{table}
\begin{center}
{\caption{The table caption is centered on the table measure. If it
extends to two lines each is centered.}\label{table1}}
\begin{tabular}{lccccccc}
\hline
\rule{0pt}{12pt}
&\multicolumn{7}{c}{Processors}\\
&1&\multicolumn{3}{c}{2}&\multicolumn{3}{c}{4}\\
\cline{2-8}
\rule{0pt}{12pt}
Window&$\Diamond$&$\Diamond$&$\Box$&$\bigtriangleup$&$\Diamond$&$\Box$&$\bigtriangleup$
\\
\hline
\\[-6pt]
\quad1&1273&110&21.79&89\%&6717&22.42&61\%\\
\quad2&2145&116&10.99&50\%&5386&10.77&19\%\\
\quad3&3014&117&41.77&89\%&7783&42.31&58\%\\
\quad4&4753&151&71.55&77\%&7477&61.97&49\%\\
\quad5&5576&148&61.60&80\%&7551&91.80&45\%
\\
\hline
\\[-6pt]
\multicolumn{8}{l}{$\Diamond$ execution time in ticks\ \
$\Box$ speed-up values\ \
$\bigtriangleup$ efficiency values}
\end{tabular}
\end{center}
\end{table}

\section{Figures}
A figure caption is set centered in 8 point (2.8 mm) medium on a
leading of 10 point (3.5 mm).  It is set under the figure, with the
word Figure and the number in bold and with a 1 pica (4.2 mm) space
separating the caption text from the figure number.

One line of space separates the figure from the caption. A one line
space separates the figure from surrounding text.

\begin{figure}
\centerline{\includegraphics[height=3.5in]{ecaif01}}
\caption{Network of transputers and the structure of individual
processes } \label{procstructfig}
\end{figure}

\section{Equations}
A display equation is numbered, using arabic numbers in parentheses.
It is centered and set with one line of space above and below to
separate it from surrounding text. The following example is a simple
single line equation:

\begin{equation}
Ax=b
\label{thesystem}
\end{equation}

The next example is a multi-line equation:

\begin{eqnarray}
(x+y)(x-y) & = & x^2-xy+xy-y^2\\
(x+y)^2    & = & x^2+2xy+y^2
\end{eqnarray}

The equal signs are aligned in a multi-line equation.

\section{Program listings}
Program listings are set in 9 point (3.15 mm) Courier on a
leading of 11 point (3.85 mm). That is to say, a non-proportional
font is used to ensure the correct alignment.

A one line space separates the program listing from surrounding text.

\begin{verbatim}
void inc(x)
int* x;
{
    *x++;
}

int y = 1;
inc(&y);
printf("%d\n",y);
\end{verbatim}

\section{Theorems}
The text of a theorem is set in 9 point (3.15 mm) italic on a
leading of 11 point (3.85 mm). The word Theorem and its number
are set in 9 point (3.15 mm) bold.

A one line space separates the theorem from surrounding text.

\begin{theorem}
Let us assume this is a valid theorem. In reality it is a piece
of text set in the theorem environment.
\end{theorem}

\section{Footnotes}
Footnotes are set in 8 point (2.8 mm) medium with leading of 8.6 point (3.1
mm), with a 1 point (0.35 mm) footnote rule to column
width\footnote{This is an example of a footnote that occurs in
the text. If the text runs to two lines the second line aligns
with the start of text in the first line.} .

\section{References}
The reference identifier in the text is set as a sequential number in square brackets. The reference entry itself is set in 8 point (2.8 mm) with a leading of 10 point (3.5 mm), and the list of references is sorted alphabetically.

\section{Sample coding}
The remainder of this paper contains examples of the specifications
detailed above and can be used for reference if required.

\section{Programming model}
\label{par}
Our algorithms were implemented using the \emph{single program,
  multiple data} model (SPMD). SPMD involves writing a single code
that will run on all the processors co-operating on a task. The data
are partitioned among the processors which know what portions of the
data they will work on \cite{kn:Golub89}.

\subsection{Structure of processes and processors}
\label{procstruct}
The grid has $P=P_{\rm{r}}\times P_{\rm{c}}$ processors, where
$P_{\rm{r}}$ is the number of rows of processors and $P_{\rm{c}}$ is
the number of columns of processors.

\subsubsection{Routing information on the grid}
\label{routing} A message may be either \emph{broadcast} or {\em
  specific}. A broadcast message originates on a processor and is
relayed through the network until it reaches all other processors. A
specific message is one that is directed to a particular target
processor.

Broadcast messages originate from a processor called \emph{central}
which is situated in the `middle' of the grid. This processor has
co-ordinates $(\lfloor P_{\rm{r}}/2 \rfloor,\lfloor P_{\rm{c}}/2
\rfloor)$.  Messages are broadcast using the \emph{row--column
  broadcast} algorithm (RCB), which uses the following strategy.  The
number of steps required to complete the RCB algorithm (i.e., until
all processors have received the broadcast value) is given by $\lfloor
P_{\rm{r}}/2\rfloor+\lfloor P_{\rm{c}}/2\rfloor$.

A specific message is routed through the processors using the
\emph{find-row--find-column} algorithm (FRFC) detailed in
\cite{kn:deCarlini91}.  The message is sent from the \emph{originator}
processor vertically until it reaches a processor sitting in the same
row as the \emph{target} processor. The message is then moved
horizontally across the processors in that row until it reaches the
target processor. An accumulation based on the recursive doubling
technique \cite[pp. 56--61]{kn:Modi88}, would require the same number
of steps as the RCB requires. If either the row or column of the
originator and target processors are the same then the message will
travel only in a horizontal or vertical direction, respectively (see
\cite{kn:Smith85}).

\section{Data partitioning}
\label{datapart}
We use \emph{data partitioning by contiguity}, defined in the
following way. To partition the data (i.e., vectors and matrices)
among the processors, we divide the set of variables
\mbox{$V=\{\,i\,\}_{i=1}^{N}$} into $P$ subsets
$\{\,W_{p}\,\}_{p=1}^{P}$ of $s=N/P$ elements each. We assume without
loss of generality that $N$ is an integer multiple of $P$. We define
each subset as $W_{p}=\{(p-1)s+j\}_{j=1}^{s}$ (see
\cite{kn:Schofield89}, \cite{kn:daCunha92a} and \cite{kn:Atkin} for
details).

Each processor $p$ is responsible for performing the computations over
the variables contained in $W_{p}$. In the case of vector operations,
each processor will hold segments of $s$ variables. The data
partitioning for operations involving matrices is discussed in Section
\ref{mv}.

\section{Linear algebra operations}
\label{blas}
\subsection{Saxpy}
\label{saxpy}
The saxpy $w=u+\alpha v$ operation, where $u$, $v$ and $w$ are vectors
and $\alpha$ is a scalar value, has the characteristic that its
computation is \emph{disjoint elementwise} with respect to $u, v$ and
$w$. This means that we can compute a saxpy without any communication
between processors; the resulting vector $w$ does not need to be
distributed among the processors. Parallelism is exploited in the
saxpy by the fact that $P$ processors will compute the same operation
with a substantially smaller amount of data. The saxpy is computed as
\begin{equation}
w_{i}=u_{i}+\alpha v_{i},\quad \forall i\in\{W_{p}\}_{p=1}^{P}
\label{distsaxpy}
\end{equation}
\subsection{Inner-product and vector 2-norm}
\label{utv}
The inner-product $\alpha=u^{T}v=\sum_{i=1}^{N}{u_{i}v_{i}}$ is an
operation that involves accumulation of data, implying a high level of
communication between all processors. The mesh topology and the
processes architecture used allowed a more efficient use of the
processors than, for instance, a ring topology, reducing the time that
processors are idle waiting for the computed inner-product value to
arrive, but the problem still remains. The use of the SPMD paradigm
also implies the global broadcast of the final computed value to all
processors.

The inner-product is computed in three distinct phases. Phase 1 is the
computation of partial sums of the form
\begin{equation}
\alpha_{p}=\sum_{\forall i\in\{W_{p}\}}{u_{i}\times v_{i}},\quad p=1,\ldots,P
\label{distutv}
\end{equation}

The accumulation phase of the inner-product using the RCA algorithm is
completed in the same number of steps as the RCB algorithm (Section
\ref{routing}). This is because of the need to relay partial values
between processors without any accumulation taking place, owing to the
connectivity of the grid topology.

The vector 2-norm $\alpha=||\,u\,||_{2}=\sqrt{u^{T}u}$ is computed
using the inner-product algorithm described above. Once the
inner-product value is received by a processor during the final
broadcast phase, it computes the square root of that value giving the
required 2-norm value.

\subsection{Matrix--vector product}
\label{mv}
For the matrix--vector product $v=Au$, we use a \emph{column
  partitioning} of $A$.  Each processor holds a set $W_{p}$ (see
Section \ref{datapart}) of $s$ columns each of $N$ elements of $A$ and
$s$ elements of $u$.  The $s$ elements of $u$ stored locally have a
one-to-one correspondence to the $s$ columns of $A$ (e.g. a processor
holding element $u_{j}$ also holds the $j$-th column of $A$). Note
that whereas we have $A$ partitioned by columns among the processors,
the matrix--vector product is to be computed by \emph{rows}.

The algorithm for computing the matrix--vector product using column
partitioning is a generalization of the inner-product algorithm
described in Section \ref{utv} (without the need for a final broadcast
phase). At a given time during the execution of the algorithm, each
one of $P-1$ processors is computing a vector $w$ of $s$ elements
containing partial sums required for the segment of the vector $v$ in
the remaining `target' processor.  After this computation is complete,
each of the $P$ processors stores a vector $w$.  The resulting segment
of the matrix--vector product vector which is to be stored in the
target processor is obtained by summing together the $P$ vectors $w$,
as described below.

Each processor other than the target processor sends its $w$ vector to
one of its neighboring processors. A processor decides whether to send
the vector in either the row or column direction to reach the target
processor based on the FRFC algorithm (see Section \ref{routing}). If
a vector passes through further processors in its route to the target
processor the $w$ vectors are accumulated. Thus the target processor
will receive at most four $w$ vectors which, when summed to its own
$w$ vector, yield the desired set of $s$ elements of $v$.

\subsection{Matrix--vector product---finite-difference approximation}
We now consider a preconditioned version of the conjugate-gradients
method \cite{kn:Golub89}. Note that we do not need to form $A$
explicitly.  This implies a very low degree of information exchange
between the processors which can be effectively exploited with
transputers, since the required values of $u$ can be exchanged
independently through each link.

The preconditioning used in our implementations is the polynomial
preconditioning (see \cite{kn:Saad85}, \cite{kn:Eisenstat81},
\cite{kn:Adams85} and \cite{kn:Johnson83}), which can be implemented
very efficiently in a parallel architecture since it is expressed as a
sequence of saxpys and matrix--vector products.

We have $l$ rows and columns in the discretization grid, which we want
to partition among a $P_{\rm{r}}\times P_{\rm{c}}$ mesh of
processors. Each processor will then carry out the computations
associated with a block of $\lfloor
l/P_{\rm{r}}\rfloor+\hbox{sign}\left( l\bmod P_{\rm{r}}\right)$ rows
and $\lfloor l/P_{\rm{c}}\rfloor+\hbox{sign}\left( l\bmod
  P_{\rm{c}}\right)$ columns of the interior points of the grid.

The matrix--vector product using the column partitioning is highly
parallel. Since there is no broadcast operation involved, as soon as a
processor on the boundary of the grid (either rows or columns) has
computed and sent a $w_{p}$ vector destined to a processor `A', it can
compute and (possibly) send a $w_{p}$ vector to processor `B', at
which time its neighboring processors may also have started computing
and sending their own $w$ vectors to processor `B'.

At a given point in the matrix--vector product computation, the
processors are computing $w$ vectors destined to processor A. When
these vectors have been accumulated in the row of that processor (step
1), the processors in the top and bottom rows compute and send the $w$
vectors for processor B, while the processors on the left and right
columns of the row of processor A send the accumulated $r$ vectors to
processor A (step 2). Processor A now stores its set of the resulting
$v$ vector (which is the accumulation of the $w$ vectors). In step 3,
the processors in the bottom row compute and send the $w$ vectors for
processor C while the processor at the left-hand end of the row of
processor B sends the accumulated $w$ vectors of that column towards
processor B. The next steps are similar to the above.

In our implementation, we exploit the geometry associated with the
regular grid of points used to approximate the PDE. A geometric
partitioning is used to match the topology and connectivity present in
the grid of transputers (Section \ref{procstruct}).

The discretization of the PDE is obtained by specifying a grid size
$l$ defining an associated grid of $N=l^2$ interior points (note that
this is the order of the linear system to be solved). With each
interior point, we associate a set of values, namely the coefficients
$C, N, S, E\,$ and $W$.

\section{Conclusion}
We have shown that an iterative method such as the preconditioned
conjugate-gradients method may be successfully parallelized by using
highly efficient parallel implementations of the linear algebra
operations involved. We have used the same approach to parallelize
other iterative methods with similar degrees of efficiency (see
\cite{kn:daCunha92a} and \cite {kn:daCunha92b}).

\ack We would like to thank the referees for their comments, which
helped improve this paper considerably

\bibliography{ecai}
\fi

\end{document}